# Interpolation consistency training for semi-supervised learning

Vikas Verma [a,b,*], Kenji Kawaguchi [c], Alex Lamb [a], Juho Kannala [b], Arno Solin [b], Yoshua Bengio [a], David Lopez-Paz [d]

[a] Montreal Institute for Learning Algorithms (MILA), Canada
[b] Aalto University, Finland
[c] Harvard University, USA
[d] Facebook AI Research, France

ABSTRACT

We introduce Interpolation Consistency Training (ICT), a simple and computation efficient algorithm for training Deep Neural Networks in the semi-supervised learning paradigm. ICT encourages the prediction at an interpolation of unlabeled points to be consistent with the interpolation of the predictions at those points. In classification problems, ICT moves the decision boundary to low-density regions of the data distribution. Our experiments show that ICT achieves state-of-the-art performance when applied to standard neural network architectures on the CIFAR-10 and SVHN benchmark datasets. Our theoretical analysis shows that ICT corresponds to a certain type of data-adaptive regularization with unlabeled points which reduces overfitting to labeled points under high confidence values.

## 1. Introduction

Deep learning achieves excellent performance in supervised learning tasks where labeled data is abundant (LeCun et al., 2015). However, labeling large amounts of data is often prohibitive due to time, financial, and expertise constraints. As machine learning permeates an increasing variety of domains, the number of applications where unlabeled data is voluminous and labels are scarce increases. For instance, recognizing documents in extinct languages, where a machine learning system has access to a few labels, produced by highly-skilled scholars (Clanuwat et al., 2018).

The goal of Semi-Supervised Learning (SSL) (Chapelle et al., 2010) is to leverage large amounts of unlabeled data to improve the performance of supervised learning over small datasets. Often, SSL algorithms use unlabeled data to learn additional structure about the input distribution. For instance, the existence of cluster structures in the input distribution could hint the separation of samples into different labels. This is often called the *cluster assumption*: if two samples belong to the same cluster in the input distribution, then they are likely to belong to the same class. The cluster assumption is equivalent to the *low-density separation assumption*: the decision boundary should lie in the low-density regions. The equivalence is easy to infer: A decision boundary which lies in a high-density region, will cut a cluster into two different classes, requiring that samples from different classes lie in the same cluster; which is the violation of the *cluster assumption*. The *low-density separation assumption* has inspired many recent *consistency-regularization* semi-supervised learning techniques, including the $\Pi$-model (Laine & Aila, 2017; Sajjadi et al., 2016), temporal ensembling (Laine & Aila, 2017), VAT (Miyato et al., 2018), and the Mean-Teacher (Tarvainen & Valpola, 2017).

*Consistency regularization* methods for semi-supervised learning enforce the low-density separation assumption by encouraging invariant prediction $f(u) = f(u + \delta)$ for perturbations $u+\delta$ of unlabeled points $u$. Such consistency and small prediction error can be satisfied simultaneously if and only if the decision boundary traverses a low-density path.

Different consistency regularization techniques vary in how they choose the unlabeled data perturbations $\delta$. One simple alternative is to use random perturbations $\delta$. However, random perturbations are inefficient in high dimensions, as only a tiny proportion of input perturbations are capable of pushing the decision boundary into low-density regions. To alleviate this issue, Virtual Adversarial Training or VAT (Miyato et al., 2018), searches for small perturbations $\delta$ that maximize the change in the prediction of the model. This involves computing the gradient of the predictor with respect to its input, which can be expensive for large neural network models.

* Corresponding author at: Aalto University, Finland.
*E-mail addresses:* vikasverma.iitm@gmail.com (V. Verma), kkawaguchi@fas.harvard.edu (K. Kawaguchi), lambalex@iro.umontreal.ca (A. Lamb), juho.kaanala@aalto.fi (J. Kannala), arno.solin@aalto.fi (A. Solin), yoshua.bengio@mila.quebec (Y. Bengio), dlp@fb.com (D. Lopez-Paz).

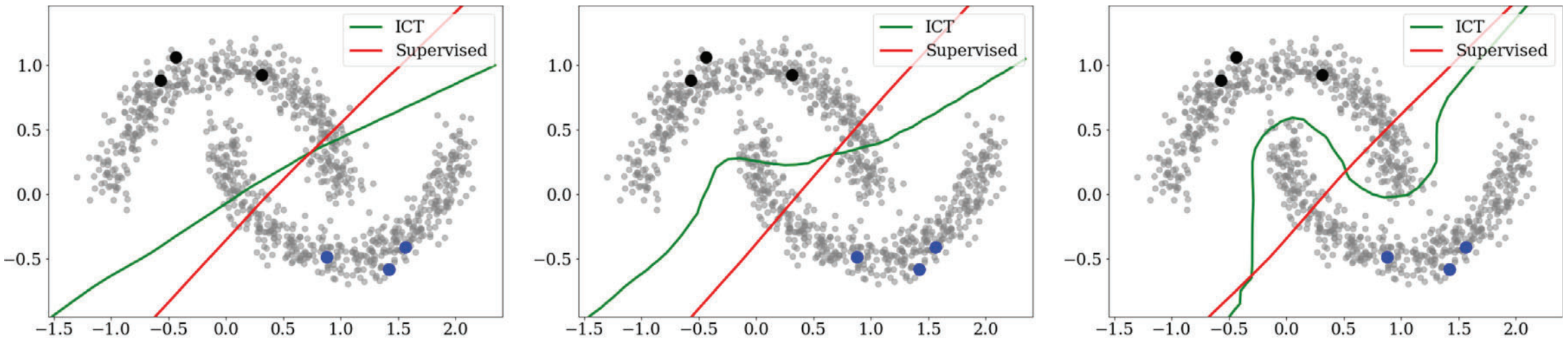

(a) After 100 updates (b) After 500 updates (c) After 1000 updates

**Fig. 1.** Interpolation Consistency Training (ICT) applied to the "two moons" dataset, when three labels per class (large dots) and a large amount of unlabeled data (small dots) are available. When compared to supervised learning (red), ICT encourages a decision boundary traversing a low-density region that would better reflect the structure of the unlabeled data. Both methods employ a multilayer perceptron with three hidden ReLU layers of twenty neurons. Best viewed in colors in the printed version.

This additional computation makes VAT (Miyato et al., 2018) and other related methods such as Park et al. (2018) less appealing in situations where unlabeled data is available in large quantities. Furthermore, recent research has shown that training with adversarial perturbations can hurt generalization performance (Nakkiran, 2019; Tsipras et al., 2018).

To overcome the above limitations, we propose the Interpolation Consistency Training (ICT), an efficient consistency regularization technique for state-of-the-art semi-supervised learning. In a nutshell, ICT regularizes semi-supervised learning by encouraging consistent predictions $f(\alpha u_1 + (1-\alpha)u_2) = \alpha f(u_1) + (1-\alpha)f(u_2)$ at interpolations $\alpha u_1 + (1-\alpha)u_2$ of unlabeled points $u_1$ and $u_2$.

Our experimental results on the benchmark datasets CIFAR10 and SVHN and neural network architectures CNN-13 (Laine & Aila, 2017; Luo et al., 2018; Miyato et al., 2018; Park et al., 2018; Tarvainen & Valpola, 2017) and WRN28-2 (Oliver et al., 2018) outperform (or are competitive with) the state-of-the-art methods. ICT is simpler and more computation efficient than several of the recent SSL algorithms, making it an appealing approach to SSL. Fig. 1 illustrates how ICT learns a decision boundary traversing a low density region in the "two moons" problem.

After the publication of an earlier version of this paper (Verma et al., 2019), similar ideas have been explored in Berthelot, Carlini et al. (2019) and Berthelot et al. (2020), with some additional techniques in comparison to that proposed in Verma et al. (2019). Although all of these methods work well in practice, the theoretical underpinning of these methods is not well understood. In this work, we additionally provide a novel theory of ICT to understand how and when ICT can succeed or fail to effectively utilize unlabeled points.

## 2. Interpolation consistency training

Given a mixup (Zhang et al., 2018) operation:

$$\mathrm{Mix}_\lambda(a, b) = \lambda \cdot a + (1-\lambda)\cdot b,$$

Interpolation Consistency Training(ICT) trains a prediction model $f_\theta$ to provide consistent predictions at interpolations of unlabeled points:

$$f_\theta(\mathrm{Mix}_\lambda(u_j, u_k)) \approx \mathrm{Mix}_\lambda(f_{\theta'}(u_j), f_{\theta'}(u_k)),$$

where $\theta'$ is a moving average of $\theta$ (Fig. 2). But, why do interpolations between unlabeled samples provide a good consistency perturbation for semi-supervised training?

To begin with, observe that the most useful samples on which the consistency regularization should be applied are the samples near the decision boundary. Adding a small perturbation $\delta$ to such low-margin unlabeled samples $u_j$ is likely to push $u_j + \delta$ over the other side of the decision boundary. This would violate the *low-density separation assumption*, making $u_j + \delta$ a good place to apply consistency regularization. These violations do not occur at high-margin unlabeled points that lie far away from the decision boundary.

Back to low-margin unlabeled points $u_j$, how can we find a perturbation $\delta$ such that $u_j$ and $u_j + \delta$ lie on opposite sides of the decision boundary? Although tempting, using random perturbations is an inefficient strategy, since the subset of directions approaching the decision boundary is a tiny fraction of the ambient space.

Instead, consider interpolations $u_j + \delta = \mathrm{Mix}_\lambda(u_j, u_k)$ towards a second randomly selected unlabeled examples $u_k$. Then, the two unlabeled samples $u_j$ and $u_k$ can either:

1. lie in the same cluster,
2. lie in different clusters but belong to the same class,
3. lie on different clusters and belong to the different classes.

Assuming the cluster assumption, the probability of (1) decreases as the number of classes increases. The probability of (2) is low if we assume that the number of clusters for each class is balanced. Finally, the probability of (3) is the highest. Then, assuming that one of $(u_j, u_k)$ lies near the decision boundary (it is a good candidate for enforcing consistency), it is likely (because of the high probability of (3)) that the interpolation towards $u_k$ points towards a region of low density, followed by the cluster of the other class. Since this is a good direction to move the decision, the interpolation is a good perturbation for consistency-based regularization.

Our exposition has argued so far that interpolations between random unlabeled samples are likely to fall in low-density regions. Thus, such interpolations are good locations where consistency-based regularization could be applied. But how should we label those interpolations? Unlike random or adversarial perturbations of single unlabeled examples $u_j$, our scheme involves two unlabeled examples $(u_j, u_k)$. Intuitively, we would like to push the decision boundary as far as possible from the class boundaries, as it is well known that decision boundaries with large margin generalize better (Shawe-Taylor et al., 1996). In the supervised learning setting, one method to achieve large-margin decision boundaries is mixup (Zhang et al., 2018). In mixup, the decision boundary is pushed far away from the class boundaries by enforcing the prediction model to change linearly in between samples. This is done by training the model $f_\theta$ to predict $\mathrm{Mix}_\lambda(y, y')$ at location $\mathrm{Mix}_\lambda(x, x')$, for random pairs of labeled samples $((x, y), (x', y'))$. Here we extend mixup to the semi-supervised learning setting by training the model $f_\theta$ to predict the "fake label" $\mathrm{Mix}_\lambda(f_\theta(u_j), f_\theta(u_k))$ at location $\mathrm{Mix}_\lambda(u_j, u_k)$. In order to follow a more conservative consistent regularization, we encourage the model $f_\theta$ to predict the fake label $\mathrm{Mix}_\lambda(f_{\theta'}(u_j), f_{\theta'}(u_k))$ at location $\mathrm{Mix}_\lambda(u_j, u_k)$, where $\theta'$ is a moving average of $\theta$, also known as a *mean-teacher* (Tarvainen & Valpola, 2017).

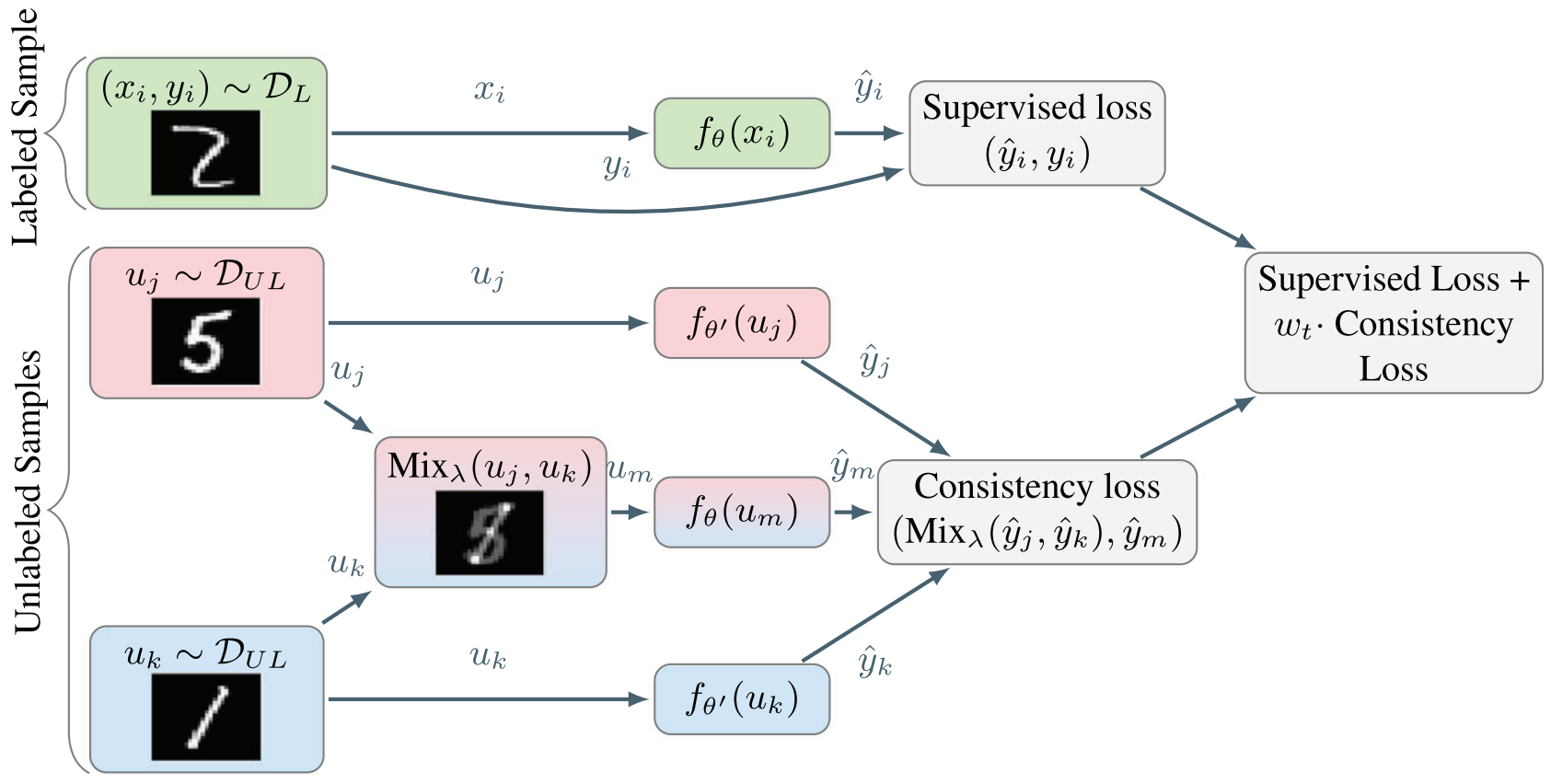

**Fig. 2.** Interpolation Consistency Training (ICT) learns a student network $f_\theta$ in a semi-supervised manner. To this end, ICT uses a mean-teacher $f_{\theta'}$, where the teacher parameters $\theta'$ are an exponential moving average of the student parameters $\theta$. During training, the student parameters $\theta$ are updated to encourage consistent predictions $f_\theta(\mathrm{Mix}_\lambda(u_j, u_k)) \approx \mathrm{Mix}_\lambda(f_{\theta'}(u_j), f_{\theta'}(u_k))$, and correct predictions for labeled examples $x_i$.

We are now ready to describe in detail the proposed Interpolation Consistency Training (ICT). Consider access to labeled samples $(x_i, y_i) \sim \mathcal{D}_L$, drawn from the joint distribution $P_{XY}(X, Y)$. Also, consider access to unlabeled samples $u_j, u_k \sim \mathcal{D}_{UL}$, drawn from the marginal distribution $P_X(X) = \frac{P_{XY}(X,Y)}{P_{Y|X}(Y|X)}$. Our learning goal is to train a model $f_\theta$, able to predict $Y$ from $X$. By using stochastic gradient descent, at each iteration $t$, update the parameters $\theta$ to minimize

$$L = L_S + w(t) \cdot L_{US}$$

where $L_S$ is the usual cross-entropy supervised learning loss over labeled samples $\mathcal{D}_L$, and $L_{US}$ is our new interpolation consistency regularization term. These two losses are computed on top of (labeled and unlabeled) minibatches, and the ramp function $w(t)$ increases the importance of the consistency regularization term $L_{US}$ after each iteration. To compute $L_{US}$, sample two minibatches of unlabeled points $u_j$ and $u_k$, and compute their fake labels $\hat{y}_j = f_{\theta'}(u_j)$ and $\hat{y}_k = f_{\theta'}(u_k)$, where $\theta'$ is an moving average of $\theta$ (Tarvainen & Valpola, 2017). Second, compute the interpolation $u_m = \mathrm{Mix}_\lambda(u_j, u_k)$, as well as the model prediction at that location, $\hat{y}_m = f_\theta(u_m)$. Third, update the parameters $\theta$ as to bring the prediction $\hat{y}_m$ closer to the interpolation of the fake labels $\mathrm{Mix}_\lambda(\hat{y}_j, \hat{y}_k)$. The discrepancy between the prediction $\hat{y}_m$ and $\mathrm{Mix}_\lambda(\hat{y}_j, \hat{y}_k)$ can be measured using any loss; in our experiments, we use the mean squared error. Following Zhang et al. (2018), on each update we sample a random $\lambda$ from $\mathrm{Beta}(\alpha, \alpha)$.

In sum, the population version of our ICT term can be written as:

$$\mathcal{L}_{US} = \mathop{\mathbb{E}}_{u_j, u_k \sim P_X} \mathop{\mathbb{E}}_{\lambda \sim \mathrm{Beta}(\alpha,\alpha)} \ell(f_\theta(\mathrm{Mix}_\lambda(u_j, u_k)), \mathrm{Mix}_\lambda(f_{\theta'}(u_j), f_{\theta'}(u_k))) \tag{1}$$

ICT is summarized in Fig. 2 and Algorithm 1.

## 3. Experiments

### *3.1. Datasets*

We follow the common practice in semi-supervised learning literature (Laine & Aila, 2017; Luo et al., 2018; Miyato et al., 2018; Park et al., 2018; Tarvainen & Valpola, 2017) and conduct experiments using the CIFAR-10, SVHN, and CIFAR-100 datasets, where only a fraction of the training data is labeled, and the remaining data is used as unlabeled data. We followed the standardized procedures laid out by Oliver et al. (2018) to ensure a fair comparison.

The CIFAR-10 dataset consists of 60 000 color images each of size 32 × 32, split between 50K training and 10K test images. This dataset has ten classes, which include images of natural objects such as cars, horses, airplanes and deer. The CIFAR-100 is similar to the CIFAR-10 dataset, except it has 100 classes with 600 images in each class. The SVHN dataset consists of 73 257 training samples and 26 032 test samples each of size 32 × 32. Each example is a close-up image of a house number (the ten classes are the digits from 0–9).

We adopt the standard data-augmentation and pre-processing scheme which has become standard practice in the semi-supervised learning literature (Athiwaratkun et al., 2019; Laine & Aila, 2017; Luo et al., 2018; Miyato et al., 2018; Sajjadi et al., 2016; Tarvainen & Valpola, 2017). More specifically, for CIFAR-10, we first zero-pad each image with 2 pixels on each side. Then, the resulting image is randomly cropped to produce a new 32 × 32 image. Next, the image is horizontally flipped with probability 0.5, followed by per-channel standardization and ZCA preprocessing. For SVHN, we zero-pad each image with 2 pixels on each side and then randomly crop the resulting image to produce a new 32 × 32 image, followed by zero-mean and unit-variance image whitening.

### *3.2. Models*

We conduct our experiments using CNN-13 and Wide-Resnet-28-2 architectures. The CNN-13 architecture has been adopted as the standard benchmark architecture in recent state-of-the-art SSL methods (Laine & Aila, 2017; Luo et al., 2018; Miyato et al., 2018; Park et al., 2018; Tarvainen & Valpola, 2017). We use its variant (i.e., without additive Gaussian noise in the input layer) as implemented in Athiwaratkun et al. (2019). We also removed the Dropout noise to isolate the improvement achieved through our method. Other SSL methods in Tables 1 and 2 use the Dropout noise, which gives them more regularizing capabilities. Despite this, our method outperforms other methods in several experimental settings.

Oliver et al. (2018) performed a systematic study using Wide-Resnet-28-2 (Zagoruyko & Komodakis, 2016), a specific residual network architecture, with extensive hyperparameter search to compare the performance of various consistency-based semi-supervised algorithms. We evaluate ICT using this same setup as a mean towards a fair comparison to these algorithms.

**Algorithm 1** The Interpolation Consistency Training (ICT) Algorithm

**Require:** $f_\theta(x)$: neural network with trainable parameters $\theta$
**Require:** $f_{\theta'}(x)$ mean teacher with $\theta'$ equal to moving average of $\theta$
**Require:** $\mathcal{D}_L(x, y)$: collection of the labeled samples
**Require:** $\mathcal{D}_{UL}(x)$: collection of the unlabeled samples
**Require:** $\alpha$: rate of moving average
**Require:** $w(t)$: ramp function for increasing the importance of consistency regularization
**Require:** $T$: total number of iterations
**Require:** $Q$: random distribution on [0,1]
**Require:** $\text{Mix}_\lambda(a, b) = \lambda a + (1 - \lambda)b$.
 **for** $t = 1, \ldots, T$ **do**
  Sample $\{(x_i, y_i)\}_{i=1}^B \sim \mathcal{D}_L(x, y)$ ▷ Sample labeled minibatch
  $L_S = \text{CrossEntropy}(\{(f_\theta(x_i), y_i)\}_{i=1}^B)$ ▷ Supervised loss (cross-entropy)
  Sample $\{u_j\}_{j=1}^U, \{u_k\}_{k=1}^U \sim \mathcal{D}_{UL}(x)$ ▷ Sample two unlabeled minibatchs
  $\{\hat{y}_j\}_{j=1}^U = \{f_{\theta'}(u_j)\}_{j=1}^U, \{\hat{y}_k\}_{k=1}^U = \{f_{\theta'}(u_k)\}_{k=1}^U$ ▷ Compute pseudo labels
  Sample $\lambda \sim Q$ ▷ sample an interpolation coefficient
  $(u_m = \text{Mix}_\lambda(u_j, u_k), \hat{y}_m = \text{Mix}_\lambda(\hat{y}_j, \hat{y}_k))$ ▷ Compute interpolation
  $L_{US} = \text{ConsistencyLoss}(\{(f_\theta(u_m), \hat{y}_m)\}_{m=1}^U)$ ▷ e.g., mean squared error
  $L = L_S + w(t) \cdot L_{US}$ ▷ Total Loss
  $g_\theta \leftarrow \nabla_\theta L$ ▷ Compute Gradients
  $\theta' = \alpha\theta' + (1 - \alpha)\theta$ ▷ Update moving average of parameters
  $\theta \leftarrow \text{Step}(\theta, g_\theta)$ ▷ e.g. SGD, Adam
 **end for**
 **return** $\theta$

**Table 1**
Error rates (%) on CIFAR-10 using CNN-13 architecture. We ran three trials for ICT.

| Model | 1000 labeled 50 000 unlabeled | 2000 labeled 50 000 unlabeled | 4000 labeled 50 000 unlabeled |
|---|---|---|---|
| Supervised | 39.95 ± 0.75 | 31.16 ± 0.66 | 21.75 ± 0.46 |
| Supervised (Mixup) | 36.48 ± 0.15 | 26.24 ± 0.46 | 19.67 ± 0.16 |
| Supervised (Manifold Mixup) | 34.58 ± 0.37 | 25.12 ± 0.52 | 18.59 ± 0.18 |
| Π model (Laine & Aila, 2017) | 31.65 ± 1.20 | 17.57 ± 0.44 | 12.36 ± 0.31 |
| TempEns (Laine & Aila, 2017) | 23.31 ± 1.01 | 15.64 ± 0.39 | 12.16 ± 0.24 |
| MT (Tarvainen & Valpola, 2017) | 21.55 ± 1.48 | 15.73 ± 0.31 | 12.31 ± 0.28 |
| VAT (Miyato et al., 2018) | – | – | 11.36 ± NA |
| VAT+Ent (Miyato et al., 2018) | – | – | 10.55 ± NA |
| VAdD (Park et al., 2018) | – | – | 11.32 ± 0.11 |
| SNTG (Luo et al., 2018) | 18.41 ± 0.52 | 13.64 ± 0.32 | 10.93 ± 0.14 |
| MT+ Fast SWA (Athiwaratkun et al., 2019) | 15.58 ± NA | 11.02 ± NA | 9.05 ± NA |
| ICT | **15.48 ± 0.78** | **9.26 ± 0.09** | **7.29 ± 0.02** |

### 3.3. Implementation details

We used the SGD with nesterov momentum optimizer for all of our experiments. For the experiments in Tables 1 and 2, we run the experiments for 400 epochs. For the experiments in Table 3, we run experiments for 600 epochs. The initial learning rate was set to 0.1 for CIFAR-10 and SVHN and 0.25 for CIFAR-100, which is then annealed using the cosine annealing technique proposed in Loshchilov and Hutter (2016) and used by Tarvainen and Valpola (2017). The momentum parameter was set to 0.9. We used an L2 regularization coefficient 0.0001 and a batch-size of 100 in our experiments.

In each experiment, we report mean and standard deviation across three independently run trials.

The consistency coefficient $w(t)$ is ramped up from its initial value 0.0 to its maximum value at one-fourth of the total number of epochs using the same sigmoid schedule of Tarvainen and Valpola (2017). We used MSE loss for computing the consistency loss following Laine and Aila (2017) and Tarvainen and Valpola (2017). We set the decay coefficient for the mean-teacher to 0.999 following Tarvainen and Valpola (2017).

We conduct hyperparameter search over the two hyperparameters introduced by our method: the maximum value of the consistency coefficient $w(t)$ (we searched over the values in {1.0, 10.0, 20.0, 50.0, 100.0}) and the parameter $\alpha$ of distribution $Beta(\alpha, \alpha)$ (we searched over the values in {0.1, 0.2, 0.5, 1.0}). We select the best hyperparameter using a validation set of 5000 and 1000 labeled samples for CIFAR-10 and SVHN respectively. This size of the validation set is the same as that used in the other methods compared in this work.

**Table 2**
Error rates (%) on SVHN using CNN-13 architecture. We ran three trials for ICT.

| Model | 250 labeled 73 257 unlabeled | 500 labeled 73 257 unlabeled | 1000 labeled 73 257 unlabeled |
|---|---|---|---|
| Supervised | 40.62 ± 0.95 | 22.93 ± 0.67 | 15.54 ± 0.61 |
| Supervised (Mixup) | 33.73 ± 1.79 | 21.08 ± 0.61 | 13.70 ± 0.47 |
| Supervised ( Manifold Mixup) | 31.75 ± 1.39 | 20.57 ± 0.63 | 13.07 ± 0.53 |
| $\Pi$ model (Laine & Aila, 2017) | 9.93 ± 1.15 | 6.65 ± 0.53 | 4.82 ± 0.17 |
| TempEns (Laine & Aila, 2017) | 12.62 ± 2.91 | 5.12 ± 0.13 | 4.42 ± 0.16 |
| MT (Tarvainen & Valpola, 2017) | 4.35 ± 0.50 | 4.18 ± 0.27 | 3.95 ± 0.19 |
| VAT (Miyato et al., 2018) | – | – | 5.42 ± NA |
| VAT+Ent (Miyato et al., 2018) | – | – | 3.86 ± NA |
| VAdD (Park et al., 2018) | – | – | 4.16 ± 0.08 |
| SNTG (Luo et al., 2018) | **4.29 ± 0.23** | **3.99 ± 0.24** | **3.86 ± 0.27** |
| ICT | 4.78 ± 0.68 | 4.23 ± 0.15 | 3.89 ± 0.04 |

**Table 3**
Results on CIFAR-10 (4000 labels) and SVHN (1000 labels) (in test error %). All results use the same standardized architecture (WideResNet-28-2). Each experiment was run for three trials. We did not conduct any hyperparameter search and used the best hyperparameters found in the experiments of Tables 1 and 2 for CIFAR-10 (4000 labels) and SVHN (1000 labels).

| SSL Approach | CIFAR10 4000 labeled 50 000 unlabeled | SVHN 1000 labeled 73 257 unlabeled |
|---|---|---|
| Supervised[a] | 20.26 ± 0.38 | 12.83 ± 0.47 |
| Mean-Teacher[a] | 15.87 ± 0.28 | 5.65 ± 0.47 |
| VAT[a] | 13.86 ± 0.27 | 5.63 ± 0.20 |
| VAT-EM[a] | 13.13 ± 0.39 | 5.35 ± 0.19 |
| ICT | **7.66 ± 0.17** | **3.53 ± 0.07** |

[a]Refers to the results reported in Oliver et al. (2018).

We note the in all our experiments with ICT, to get the supervised loss, we perform the interpolation of labeled sample pair and their corresponding labels (as in *mixup* Zhang et al., 2018). To make sure, that the improvements from ICT are not only because of the supervised *mixup* loss, we provide the direct comparison of ICT against supervised *mixup* and *Manifold Mixup* training in Tables 1 and 2.

### 3.4. Results

We provide the results for CIFAR10 and SVHN datasets using CNN-13 architecture in Tables 1 and 2, respectively.

To justify the use of a SSL algorithm, one must compare its performance against the state-of-the-art supervised learning algorithm (Oliver et al., 2018). To this end, we compare our method against two state-of-the-art supervised learning algorithms (Verma et al., 2018; Zhang et al., 2018), denoted as Supervised(Mixup) and Supervised(Manifold Mixup), respectively in Tables 1 and 2. ICT method passes this test with a wide margin, often resulting in a two-fold reduction in the test error in the case of CIFAR10 (Table 1) and a four-fold reduction in the case of SVHN (Table 2)

Furthermore, in Table 1, we see that ICT improves the test error of other strong SSL methods. For example, in the case of 4000 labeled samples, it improves the test error of best-reported method by $\sim$ 25%. The best values of the hyperparameter max-consistency coefficient for 1000, 2000 and 4000 labels experiments were found to be 10.0, 100.0 and 100.0 respectively and the best values of the hyperparameter $\alpha$ for 1000, 2000 and 4000 labels experiments were found to be 0.2, 1.0 and 1.0 respectively. In general, we observed that for less number of labeled data, lower values of max-consistency coefficient and $\alpha$ obtained better validation errors.

For SVHN, the test errors obtained by ICT are competitive with other state-of-the-art SSL methods (Table 2). The best values of the hyperparameters max-consistency coefficient and $\alpha$ were found to be 100 and 0.1 respectively, for all the ICT results reported in Table 2.

Oliver et al. (2018) performed extensive hyperparameter search for various consistency regularization SSL algorithms using the WRN-28-2 and they report the best test errors found for each of these algorithms. For a fair comparison of ICT against these SSL algorithms, we conduct experiments on WRN-28-2 architecture. The results are shown in Table 3. ICT achieves improvement over other methods both for the CIFAR10 and SVHN datasets.

We note that unlike other SSL methods of Tables 1–3, we do not use Dropout regularizer in our implementation of CNN-13 and WRN-28-2. Using Dropout along with the ICT may further reduce the test error.

### 3.5. Ablation study

- Effect of not using the mean-teacher in ICT: We note that $\Pi$-model, VAT and VAdD methods in Tables 1 and 2 do not use a mean-teacher to make predictions on the unlabeled data. Although the mean-teacher (Tarvainen & Valpola, 2017) used in ICT does not incur any significant computation cost, one might argue that a more direct comparison with $\Pi$-model, VAT and VAdD methods requires not using a mean-teacher. To this end, we conduct an experiment on the CIFAR10 dataset, without the mean-teacher in ICT, i.e. the prediction on the unlabeled data comes from the network $f_\theta(x)$ instead of the mean-teacher network $f_{\theta'}(x)$ in Eq. (1). We obtain test errors of $19.56 \pm 0.56\%$, $14.35 \pm 0.15\%$ and $11.19 \pm 0.14\%$ for 1000, 2000, 4000 labeled samples respectively (we did not conduct any hyperparameter search for these experiments and used the best hyperparameters found in the ICT experiments of Table 1). This shows that even without a mean-teacher, ICT has major an advantage over methods such as VAT (Miyato et al., 2018) and VAdD (Park et al., 2018) that it does not require an additional gradient computation yet performs on the same level of the test error.
- Effect of not having the *mixup* supervised loss: In Section 3.3, we noted that to get the supervised loss, we perform the interpolation of labeled sample pair and their corresponding labels (*mixup* supervised loss as in Zhang et al. (2018)). Will the performance of ICT be significantly reduced by not having the *mixup* supervised loss? We conducted experiments with ICT on both CIFAR10 and SVHN with the *vanilla* supervised loss. For CIFAR10, we obtained test errors of 14.86 ± 0.39, 9.02 ± 0.12 and 8.23 ± 0.22 for 1000, 2000 and 4000 labeled samples respectively. We did not conduct any hyperparameter search and used the best values of hyperparameters (max-consistency coefficient and $\alpha$) found in the experiments of Table 1. We observe that in the case of 1000 and 2000 labeled samples, there is no increase in the test error (w.r.t having the *mixup* supervised loss), whereas in the case of 4000 labels, the test error increases by approximately 1% . This suggests that, in the low labeled data regimes, not having the *mixup* supervised loss in the ICT does not incur any significant increase in the test error.

**Table 4**
Results on CIFAR-100 with 100 labels per class using the CNN-13 architecture. We ran three trials for ICT. ICT was implemented with the ReLU activation and the softplus activation function.

| Model | Test error (%) |
|---|---|
| Supervised-only[a] | 44.56 ± 0.30 |
| *Π* model[a] | 39.19 ± 0.36 |
| TempEns[a] | 38.65 ± 0.51 |
| ICT with ReLU activation | 35.54 ± 0.26 |
| ICT with softplus activation | **34.99** ± **0.32** |

[a]Refers to the results reported in Laine and Aila (2017).

### 3.6. Results with a larger number of classes

The main experiments above are conducted on CIFAR-10 and SVHN. Each of these datasets has 10 classes for the classification task. Accordingly, we conducted additional experiments using CIFAR-100, and reported the results with the CNN-13 architecture in Table 4. Following the previous work on semi-supervised learning with CIFAR-100 and CNN-13 (Laine & Aila, 2017), we used 100 labeled points per class in this experiment. We did not conduct any hyperparameter search and used the best hyperparameters found in the experiments of CIFAR-10 as reported in Section 3.4. As can be seen in Table 4, ICT with the ReLU activation function outperformed the previous methods as expected from the above study with the CIFAR-10 and SVHN datasets. Moreover, ICT worked the best after replacing the ReLU activation function with the softplus activation function (Dugas et al., 2001): $\phi(z) = \ln(1 + \exp(\kappa z))/\kappa$ with $\kappa = 100$. Here, the value of $\kappa$ was set without any hyperparameter search and simply taken from the previous study for supervised learning with a different architecture (pre-activation ResNet with 18 layers) (Kawaguchi & Sun, 2021).

## 4. Theoretical analysis

In this section, we establish mathematical properties of ICT for binary classification with $f_\theta(u) \in \mathbb{R}$. We begin in Section 4.1 with additional notation, an introduction of real analytic functions, and a property of the Euclidian norm of the Kronecker product. Using the real analyticity and the property of the Kronecker product, we show in Section 4.2 that ICT regularizes higher-order derivatives. We conclude in Section 4.3 by showing how ICT can reduce overfitting and lead to better generalization behaviors than those without ICT.

### 4.1. Preliminaries

In our experiments, we did not use mean-teacher models $f'_\theta$ in the ablation study for CIFAR-10 and the two moon dataset in all of Figs. 1 and 3–7. Those experiments consistently show the advantages of ICT even without the mean-teacher models. Indeed, the mean teacher is not a necessary mechanism of consistency-regularization in general (e.g., see equation (1) of Han et al., 2021). Thus, to understand and disentangle the essential mechanisms of ICT, we consider the same setting as those experiments: i.e., this section focuses on the mean square loss for the unlabeled data without the mean teacher as

$$L_{US} := \mathop{\mathbb{E}}_{u,u' \sim P_X} \mathop{\mathbb{E}}_{\lambda \sim \text{Beta}(\alpha,\alpha)} \ell(f_\theta(\text{Mix}_\lambda(u, u')), \text{Mix}_\lambda(f_\theta(u), f_\theta(u'))), \quad (2)$$

where $u \in \mathcal{X} \subseteq \mathbb{R}^d$ and $\ell(a, b) = (a - b)^2$. For example, if $P(X)$ is an empirical measure on the finite unlabeled data points, we can write $L_{US}$ by

$$L_{US} = \frac{1}{n^2} \sum_{i=1}^{n} \sum_{j=1}^{n} \mathop{\mathbb{E}}_{\lambda \sim \text{Beta}(\alpha,\alpha)} \ell(f_\theta(\text{Mix}_\lambda(u_i, u_j)) - \text{Mix}_\lambda(f_\theta(u_i), f_\theta(u_j))), \quad (3)$$

because $L_{US} := \mathbb{E}_{u,u' \sim P_X} g(u, u') = \iint g(u, u') dP_X(u) dP_X(u') = \frac{1}{n^2} \sum_{i=1}^{n} \sum_{j=1}^{n} g(u_i, u_j)$ for any measurable function $g$, where the last equality used the empirical measure $P_X = \frac{1}{n} \sum_{i=1}^{n} \delta_{u_i}$ with the Dirac measures $\delta_{u_i}$. We consider the function $f_\theta$ in the form of

$$f_\theta(u) = \sigma(h_\theta(u)), \quad (4)$$

where $\sigma(a) = \frac{1}{1+e^{-a}}$ is the sigmoid function and $h_\theta(u)$ represents the pre-activation output of the last layer of a deep neural network.

The theory developed in this section requires the function $f_\theta$ to be real analytic, which is satisfied for a large class of deep neural networks used in practice. Since a composition of real analytic functions is real analytic, we only need to require that each operation in each layer satisfies the real analyticity. The convolution, affine map, skip connection, batch normalization and average pooling are all real analytic functions. Therefore, the composition of these operations preserve real analyticity. Furthermore, many common activation functions are real analytic. For example, sigmoid $\sigma$, hyperbolic tangents and softplus activations $\phi(z) = \ln(1+\exp(\kappa z))/\kappa$ are all real analytic functions, with any hyperparameter $\kappa > 0$. Here, the softplus activation can approximate the ReLU activation *for any desired accuracy* as

$$\phi(x) \to \text{relu}(x) \text{ as } \kappa \to \infty,$$

where relu represents the ReLU activation. Indeed, the ReLU activation and the softplus activation behave similarly to each other in numerical experiments with a large value of $\kappa$ such as $\kappa = 100$, which is demonstrated in the previous studies (e.g., see Kawaguchi & Sun, 2021). As a further demonstration of this fact, we used the softplus activation in Fig. 3 (and Fig. 4) and the ReLU activation in Fig. 7 (and Fig. 5): we can see that with both activation functions, ICT works well and the numerical results are consistent with our theoretical predictions. This is expected because the softplus activation function can approximate the ReLU function arbitrarily well by varying the value of $\kappa \in (0, \infty)$ and our theoretical analyses hold for any $\kappa \in (0, \infty)$. Moreover, Table 4 shows that ICT with the softplus activation works similarly to and can outperform ICT with the ReLU activation. Overall, the requirement on the real analyticity of the function $f_\theta$ is easily satisfied in practice without degrading practical performances.

We close this subsection by introducing additional notation and proving a theoretical property of the Kronecker product. Let $\hat{S} = (u_i)_{i=1}^{n}$ be a dataset with unlabeled data points and $S = ((x_i, y_i))_{i=1}^{m}$ be a dataset with labeled points. Here, $\hat{S}$ contains unlabeled data points only and is not necessarily the whole dataset used for the unsupervised loss; e.g., one can use the concatenation of $\hat{S}$ and $(x_i)_{i=1}^{n}$ for the unsupervised loss. Accordingly, $\hat{S}$ and $S$ are independent of each other. Define $\hat{\mathcal{R}}_m(\mathcal{F})$ to be the empirical Rademacher complexity of the set of functions $\mathcal{F}$, whereas $\mathcal{R}_m(\mathcal{F})$ is the Rademacher complexity of the set $\mathcal{F}$. We adopt the standard convention that $\min_{a\in\varnothing}\Psi(a) = \infty$ for any function $\Psi$ with the empty set $\varnothing$. We define the $k$th order tensor $\partial^k f_\theta(u) \in \mathbb{R}^{d\times d\times\cdots\times d}$ by

$$\partial^k f_\theta(u)_{t_1 t_2\cdots t_k} = \frac{\partial^k}{\partial u_{t_1}\partial u_{t_2}\cdots\partial u_{t_k}} f_\theta(u). \tag{5}$$

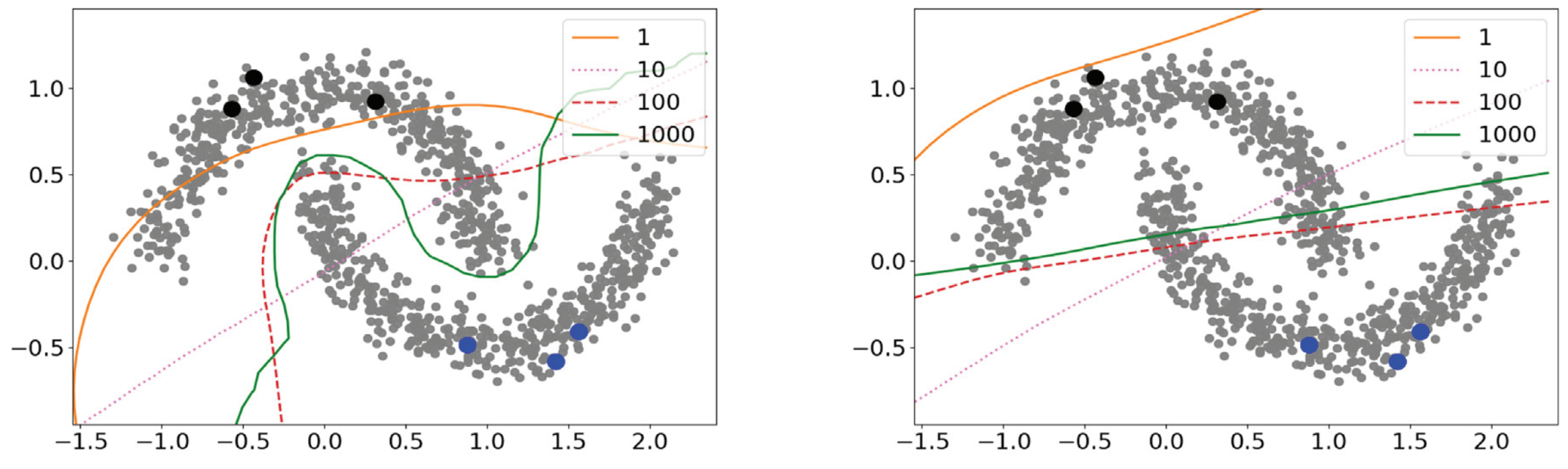

(a) High confidence value = 0.4817

(b) Low confidence value = 0.2961

**Fig. 3.** Numerical validation of the theoretical prediction that ICT performs well when the confidence value $\frac{1}{n}\sum_{i=1}^{n}|\frac{1}{2}-f_\theta(u_i)|$ is high, because that is when ICT acts as a regularizer on directional derivatives of all orders. Each line in each subplot shows the decision boundary of the predictor $f_\theta$ (i.e., $\{u : f_\theta(u)\} = \frac{1}{2}$) after each update of 1, 10, 100, and 1000. Best viewed in colors in the printed version.

For example, $\partial^1 f_\theta(u)$ is the gradient of $f_\theta$ evaluated at $u$, and $\partial^2 f_\theta(u)$ is the Hessian of $f_\theta$ evaluated at $u$. For any $k$th order tensor $\partial^k f_\theta(b) \in \mathbb{R}^{d\times d\times\cdots\times d}$, we define the vectorization of the tensor by $\mathrm{vec}[\partial^k f_\theta(b)] \in \mathbb{R}^{d^k}$. For an vector $a \in \mathbb{R}^d$, we define $a^{\otimes k} = a \otimes a \otimes \cdots \otimes a \in \mathbb{R}^{d^k}$ where $\otimes$ represents the Kronecker product. The following lemma proves that the Euclidean norm of the Kronecker products $a^{\otimes k}$ is the Euclidean norm of the vector $a$ to the $k$th power.

**Lemma 1.** *Let $d \in \mathbb{N}^+$ and $a \in \mathbb{R}^d$. Then, for any $k \in \mathbb{N}^+$,*

$$\|a^{\otimes k}\|_2 = \|a\|_2^k$$

**Proof.** We prove this statement by induction over $k \in \mathbb{N}^+$. For the base case with $k = 1$,

$$\|a^{\otimes 1}\|_2 = \|a\|_2 = \|a\|_2^1,$$

as desired. For the inductive step, we show the statement to hold for $k+1$:

$$\|a^{\otimes k+1}\|_2^2 = \|a^{\otimes k}\otimes a\|_2^2 = \left\|\begin{bmatrix}(a^{\otimes k})_1 a\\ \vdots \\ (a^{\otimes k})_{d^k} a\end{bmatrix}\right\|_2^2 = \sum_{i=1}^{d^k}((a^{\otimes k})_i)^2\|a\|_2^2$$
$$= \|a\|_2^2\sum_{i=1}^{d^k}((a^{\otimes k})_i)^2$$
$$= \|a\|_2^2\|a^{\otimes k}\|_2^2.$$

Here, the inductive hypothesis of $\|a^{\otimes k}\|_2 = \|a\|_2^k$ implies that $\|a^{\otimes k}\|_2^2 = \|a\|_2^{2k}$, and thus

$$\|a^{\otimes k+1}\|_2^2 = \|a\|_2^2\|a\|_2^{2k} = \|a\|_2^{2(k+1)}.$$

This implies that

$$\|a^{\otimes k+1}\|_2 = \|a\|_2^{k+1},$$

which completes the inductive step. □

### 4.2. Understanding ICT as regularizer on higher-order derivatives

Using the real analyticity and the Kronecker product, we show how ICT can act as a regularizer on higher-order derivatives. More concretely, Theorem 1 states that for any $K \in \mathbb{N}^+$, the ICT loss can be written as

$$\ell(f_\theta(\mathrm{Mix}_\lambda(u,u')), \mathrm{Mix}_\lambda(f_\theta(u), f_\theta(u'))) \tag{6}$$
$$= \left(\sum_{k=2}^{K}\frac{(\hat\lambda-\hat\lambda^k)}{k!}\mathrm{vec}[\partial^k f_\theta(u)]^\top(u'-u)^{\otimes k} + O(\|u'-u\|_2^K)\right)^2,$$

where $O(\|u'-u\|_2^K) \to 0$ as $K \to \infty$ if we normalize the input so that $\|u'-u\|_2 < 1$. Here, Theorem 1 holds for any $u, u' \in \mathbb{R}^d$: for example, we can replace $u$ by any $u_i$ or $x_i$, where $x_i$ represents the input part of the labeled dataset $((x_i, y_i))_{i=1}^n$.

**Theorem 1.** *Let $u, u' \in \mathbb{R}^d$ and $f_\theta$ be real analytic, and define $\Delta = u' - u$. Then, for any $K \in \mathbb{N}^+$, there exists a pair $(\zeta, \zeta') \in [0, \hat\lambda] \times [0, 1]$ such that*

$$\ell(f_\theta(\mathrm{Mix}_\lambda(u,u')), \mathrm{Mix}_\lambda(f_\theta(u), f_\theta(u')))$$
$$= \left(\sum_{k=2}^{K}\frac{(\hat\lambda-\hat\lambda^k)}{k!}\mathrm{vec}[\partial^k f_\theta(u)]^\top\Delta^{\otimes k} + E_\theta(u,u',\lambda)\right)^2, \tag{7}$$

*where $\hat\lambda = 1-\lambda$, $E_\theta(u,u',\lambda) = O(\|\Delta\|_2^K)$ and*

$$E_\theta(u,u',\lambda) = \frac{1}{K!}((1-\zeta')^K\,\mathrm{vec}[\partial^{K+1}f_\theta(u+\zeta'\Delta)]$$
$$-\hat\lambda(\hat\lambda-\zeta)^K\,\mathrm{vec}[\partial^{K+1}f_\theta(u+\zeta\Delta)])^\top\Delta^{\otimes K}.$$

**Proof.** Let $(u, u')$ be an arbitrary pair of unlabeled data points. We define the function $\varphi$ by

$$\varphi(a) = f_\theta(u + a(u'-u)).$$

Since $f_\theta$ is real analytic and $a \mapsto u + a(u'-u)$ is real analytic, their composition $\varphi$ is also real analytic. We first observe that

$$\mathrm{Mix}_\lambda(a,b) = \lambda a + (1-\lambda)b = a + (1-\lambda)(b-a) = a + \hat\lambda(b-a). \tag{8}$$

Using Eq. (8),

$$f_\theta(\mathrm{Mix}_\lambda(u,u')) = f_\theta(u + \hat\lambda(u'-u)) = \varphi(\hat\lambda).$$

Using Taylor's theorem with the Cauchy remainder, we have the following: for any $K \in \mathbb{N}^+$, there exists $\zeta \in [0, \hat{\lambda}]$ such that

$$f_\theta(\text{Mix}_\lambda(u, u')) = \varphi(\hat{\lambda}) = \varphi(0) + \sum_{k=1}^{K} \frac{\hat{\lambda}^k}{k!}\varphi^{(k)}(0) + \frac{\hat{\lambda}(\hat{\lambda}-\zeta)^K}{K!}\varphi^{(K+1)}(\zeta),$$

where $\varphi^{(k)}$ is the $k$th order derivative of $\varphi$. Since $\varphi(0) = f_\theta(u)$, this implies that

$$f_\theta(\text{Mix}_\lambda(u, u')) = f_\theta(u) + \sum_{k=1}^{K} \frac{\hat{\lambda}^k}{k!}\varphi^{(k)}(0) + \frac{\hat{\lambda}(\hat{\lambda}-\zeta)^K}{K!}\varphi^{(K+1)}(\zeta). \quad (9)$$

On the other hand, using Eq. (8),

$$\text{Mix}_\lambda(f_\theta(u), f_\theta(u')) = f_\theta(u) + \hat{\lambda}(f_\theta(u') - f_\theta(u)). \quad (10)$$

Using Taylor's theorem with the Cauchy remainder, we have the following: for any $K \in \mathbb{N}^+$, there exists $\zeta' \in [0, 1]$ such that

$$f_\theta(u') = \varphi(1) = \varphi(0) + \sum_{k=1}^{K} \frac{1}{k!}\varphi^{(k)}(0) + \frac{(1-\zeta')^K}{K!}\varphi^{(K+1)}(\zeta').$$

Since $\varphi(0) = f_\theta(u)$, plugging this formula of $f_\theta(u')$ into Eq. (10) yields

$$\text{Mix}_\lambda(f_\theta(u), f_\theta(u')) = f_\theta(u) + \hat{\lambda}\left(\sum_{k=1}^{K} \frac{1}{k!}\varphi^{(k)}(0) + \frac{(1-\zeta')^K}{K!}\varphi^{(K+1)}(\zeta')\right). \quad (11)$$

Using Eqs. (9) and (11),

$$\begin{aligned}
&\ell(f_\theta(\text{Mix}_\lambda(u, u')), \text{Mix}_\lambda(f_\theta(u), f_\theta(u'))) \\
&= \left(\sum_{k=1}^{K} \frac{\hat{\lambda}^k}{k!}\varphi^{(k)}(0) + \frac{\hat{\lambda}(\hat{\lambda}-\zeta)^K}{K!}\varphi^{(K+1)}(\zeta) - \hat{\lambda}\left(\sum_{k=1}^{K} \frac{1}{k!}\varphi^{(k)}(0) + \frac{(1-\zeta')^K}{K!}\varphi^{(K+1)}(\zeta')\right)\right)^2 \\
&= \left(\sum_{k=1}^{K} \frac{(\hat{\lambda}^k - \hat{\lambda})}{k!}\varphi^{(k)}(0) + \frac{1}{K!}(\hat{\lambda}(\hat{\lambda}-\zeta)^K\varphi^{(K+1)}(\zeta) - (1-\zeta')^K\varphi^{(K+1)}(\zeta'))\right)^2
\end{aligned}$$

Since $\hat{\lambda}^k - \hat{\lambda} = 0$ when $k = 1$, this implies that

$$\begin{aligned}
&\ell(f_\theta(\text{Mix}_\lambda(u, u')), \text{Mix}_\lambda(f_\theta(u), f_\theta(u'))) \quad (12)\\
&= \left(\sum_{k=2}^{K} \frac{(\hat{\lambda}^k - \hat{\lambda})}{k!}\varphi^{(k)}(0) + \frac{1}{K!}(\hat{\lambda}(\hat{\lambda}-\zeta)^K\varphi^{(K+1)}(\zeta) - (1-\zeta')^K\varphi^{(K+1)}(\zeta'))\right)^2.
\end{aligned}$$

We now derive the formula of $\varphi^{(k)}(a)$. By the chain rule, with $\Delta = u' - u$ and $b = u + a\Delta$, we have that

$$\varphi^{(1)}(a) = \frac{\partial f_\theta(u + a\Delta)}{\partial a} = \sum_{t_1=1}^{d} \frac{\partial f_\theta(b)}{\partial b_{t_1}}\frac{\partial b_{t_1}}{\partial a} = \sum_{t_1=1}^{d} \frac{\partial f_\theta(b)}{\partial b_{t_1}}\Delta_{t_1}$$

$$\varphi^{(2)}(a) = \frac{\partial}{\partial a}\frac{\partial f_\theta(u + a\Delta)}{\partial a} = \sum_{t_1=1}^{d} \frac{\partial}{\partial a}\frac{\partial f_\theta(b)}{\partial b_{t_1}}\Delta_{t_1} = \sum_{t_1=1}^{d}\sum_{t_2=1}^{d} \frac{\partial^2 f_\theta(b)}{\partial b_{t_1}\partial b_{t_2}}\Delta_{t_2}\Delta_{t_1}$$

Based on this process, we consider the following formula of $\varphi^{(k)}(a)$ as a candidate to be proven by induction over $k \in \mathbb{N}^+$:

$$\sum_{t_1=1}^{d}\sum_{t_2=1}^{d}\cdots\sum_{t_k=1}^{d} \frac{\partial^k f_\theta(b)}{\partial b_{t_1}\partial b_{t_2}\cdots\partial b_{t_k}}\Delta_{t_1}\Delta_{t_2}\cdots\Delta_{t_k}.$$

For the base case, we have already shown that $\varphi^{(1)}(a) = \sum_{t_1=1}^{d} \frac{\partial f_\theta(b)}{\partial b_{t_1}}\Delta_{t_1}$ as desired. For the inductive step, by using the inductive hypothesis,

$$\begin{aligned}
\varphi^{(k+1)}(a) &= \frac{\partial}{\partial a}\sum_{t_1=1}^{d}\sum_{t_2=1}^{d}\cdots\sum_{t_k=1}^{d} \frac{\partial^k f_\theta(b)}{\partial b_{t_1}\partial b_{t_2}\cdots\partial b_{t_k}}\Delta_{t_1}\Delta_{t_2}\cdots\Delta_{t_k} \\
&= \sum_{t_1=1}^{d}\sum_{t_2=1}^{d}\cdots\sum_{t_k=1}^{d}\sum_{t_{k+1}=1}^{d} \frac{\partial^{k+1} f_\theta(b)}{\partial b_{t_1}\partial b_{t_2}\cdots\partial b_{t_k}\partial b_{t_{k+1}}}\Delta_{t_1}\Delta_{t_2}\cdots\Delta_{t_k}\Delta_{t_{k+1}}
\end{aligned}$$

as desired. Therefore, we have proven that for any $k \in \mathbb{N}^+$,

$$\varphi^{(k)}(a) = \sum_{t_1=1}^{d}\sum_{t_2=1}^{d}\cdots\sum_{t_k=1}^{d} \frac{\partial^k f_\theta(b)}{\partial b_{t_1}\partial b_{t_2}\cdots\partial b_{t_k}}\Delta_{t_1}\Delta_{t_2}\cdots\Delta_{t_k}. \quad (13)$$

Then, by using the vectorization of the tensor $\text{vec}[\partial^k f_\theta(b)] \in \mathbb{R}^{d^k}$, we can rewrite Eq. (13) as

$$\varphi^{(k)}(a) = \text{vec}[\partial^k f_\theta(u + a\Delta)]^\top \Delta^{\otimes k}, \quad (14)$$

where $\Delta^{\otimes k} = \Delta \otimes \Delta \otimes \cdots \otimes \Delta \in \mathbb{R}^{d^k}$ and $\Delta = u' - u$. By combining Eqs. (12) and (14),

$$\begin{aligned}
&\ell(f_\theta(\text{Mix}_\lambda(u, u')), \text{Mix}_\lambda(f_\theta(u), f_\theta(u'))) \\
&= \left(\sum_{k=2}^{K} \frac{(\hat{\lambda}^k - \hat{\lambda})}{k!}\varphi^{(k)}(0) + \frac{1}{K!}(\hat{\lambda}(\hat{\lambda}-\zeta)^K\varphi^{(K+1)}(\zeta) - (1-\zeta')^K\varphi^{(K+1)}(\zeta'))\right)^2 \\
&= \left(\sum_{k=2}^{K} \frac{(\hat{\lambda} - \hat{\lambda}^k)}{k!}\text{vec}[\partial^k f_\theta(u)]^\top \Delta^{\otimes k} + E_\theta(u, u', \lambda)\right)^2
\end{aligned}$$

By using the Cauchy–Schwarz inequality,

$$\begin{aligned}
&|E_\theta(u, u', \lambda)| \\
&= \left|\frac{1}{K!}((1-\zeta')^K \text{vec}[\partial^{K+1} f_\theta(u + \zeta'\Delta)] - \hat{\lambda}(\hat{\lambda}-\zeta)^K \text{vec}[\partial^{K+1} f_\theta(u + \zeta\Delta)])^\top \Delta^{\otimes K}\right| \\
&\le \frac{\|\Delta^{\otimes K}\|_2}{K!}\left\|((1-\zeta')^K \text{vec}[\partial^{K+1} f_\theta(u + \zeta'\Delta)] - \hat{\lambda}(\hat{\lambda}-\zeta)^K \text{vec}[\partial^{K+1} f_\theta(u + \zeta\Delta)])\right\|_2 \\
&= \frac{\|\Delta\|_2^K}{K!}\left\|((1-\zeta')^K \text{vec}[\partial^{K+1} f_\theta(u + \zeta'\Delta)] - \hat{\lambda}(\hat{\lambda}-\zeta)^K \text{vec}[\partial^{K+1} f_\theta(u + \zeta\Delta)])\right\|_2
\end{aligned}$$

where the last line follows from Lemma 1. □

Theorem 1 shows that ICT acts as a regularizer on derivatives of all orders *when* the confidence value $\frac{1}{n}\sum_{i=1}^{n}|\frac{1}{2} - f_\theta(u_i)|$ of the prediction on unlabeled points $(u_i)_{i=1}^n$ is high.[1] This is because

[1] Here, the confidence value at an unlabeled point $u_i$ is defined by $|\frac{1}{2} - f_\theta(u_i)|$, because $f_\theta(u_i) \in (0, 1)$ is the output of the sigmoid function at the last layer of

increasing confidence in ICT tends to decrease the norm of the first-order derivatives due to the following observation. By the chain rule, the first-order derivatives can be written as

$$\partial f_\theta(u) = \partial\sigma(h_\theta(u))\partial h_\theta(u). \tag{15}$$

Moreover, since $\partial\sigma(a) = \frac{e^{-a}}{(1+e^{-a})^2} = \frac{e^{a}}{(1+e^{a})^2}$ and $\partial\sigma(h_\theta(u)) \geq 0$, we have

$$\begin{aligned}\|\partial f_\theta(u)\| &= |\partial\sigma(h_\theta(u))|\|\partial h_\theta(u)\| = \partial\sigma(h_\theta(u))\|\partial h_\theta(u)\| \\ &= \partial\sigma(|h_\theta(u)|)\|\partial h_\theta(u)\|. \end{aligned}\tag{16}$$

Here, $\partial\sigma(|h_\theta(u)|)$ is maximized when $|h_\theta(u)| = 0$ and exponentially decreases towards 0 as $|h_\theta(u)|$ increases or equivalently as the confidence of the prediction increases. In other words, $\partial\sigma(|h_\theta(u)|)$ is maximized when we have lowest confidence (i.e., $f_\theta(u) = 1/2$), and $\partial\sigma(|h_\theta(u)|)$ exponentially decreases towards 0 as we increase the confidence (i.e., as $f_\theta(u)$ moves towards 0 or 1 from 1/2). Therefore, as long as $\|\partial h_\theta(u)\|$ is bounded (or increase slower than the exponential rate), increasing the confidence of the prediction can implicitly minimize the norm of the first-order derivative $\|\partial f_\theta(u)\|$. Here, the weight decay also tends to bound $\|\partial h_\theta(u)\|$ since $\partial h_\theta(u)$ contains the products of the weight matrices for the deep networks by the chain rule.

Theorem 1 along this observation suggests that *ICT works well when the confidence on unlabeled data is high (which can be enforced by using pseudo-labels), because that is when ICT acts as a regularizer on derivatives of all orders*. To confirm this theoretical prediction, we conducted numerical simulations with the "two moons" dataset by intentionally decreasing confidence values on the unlabeled data points. Fig. 3 shows the results of this experiment with the high confidence case and the low confidence case. Here, the confidence value is defined by $\frac{1}{n}\sum_{i=1}^{n}|\frac{1}{2} - f_\theta(u_i)|$ for the unlabeled dataset $(u_i)_{i=1}^n$, which is intentionally reduced in Fig. 3(b). We used the same settings as those in Fig. 1, except that we use the softplus activation function $\phi(z) = \ln(1+\exp(\kappa z))/\kappa$ with $\kappa = 100$ and we did not use mixup for the supervised loss, in order to understand the essential mechanism of ICT (whereas we used mixup for the supervised loss in Fig. 1). As can be seen in Fig. 3, the numerical results are consistent with our theoretical prediction. The qualitatively same behaviors were also observed with the ReLU activation as shown in Appendix B.3.

### 4.3. On overfitting

In the previous subsection, we have shown that ICT acts as a regularizer on the derivatives of all orders at unlabeled data points. In this subsection, we provide a theoretical explanation regarding how regularizing the derivatives of all orders at *unlabeled* points help reducing overfitting at *labeled* points.

We first recall the important lemma, Lemma 2, that bounds a possible degree of overfitting at labeled points with the Rademacher complexity of a hypothesis space (Bartlett & Mendelson, 2002; Mohri et al., 2012). Since our notion of a hypothesis space differs from standard ones, we include a proof for the sake of completeness. The proof is based on an argument in Bartlett and Mendelson (2002). A key observation in Lemma 2 is that we can define a hypothesis space based on the unlabeled dataset $\hat{S} = (u_i)_{i=1}^n$, which is to be used later to relate the regularization at unlabeled points to the degree of overfitting at labeled points.

a deep neural network. With $f_\theta(u_i) = \frac{1}{2}$, we have the minimum confidence, where the confidence value $|\frac{1}{2} - f_\theta(u_i)|$ is zero as desired. The confidence value is maximized towards 0.5 when we let $f_\theta(u_i) \to 0$ or 1.

**Lemma 2.** *Let $\mathcal{F}_{\hat{S}}$ be a set of maps $x \mapsto f(x)$ that depends on an unlabeled dataset $\hat{S}$. Let $q \mapsto \ell(q, y)$ be a C-uniformly bounded function for any $q \in \{f(x) : f \in \mathcal{F}_{\hat{S}}, x \in \mathcal{X}\}$ and $y \in \mathcal{Y}$. Then, for any $\delta > 0$, with probability at least $1-\delta$ over an i.i.d. draw of m samples $S = ((x_i, y_i))_{i=1}^m$, the following holds: for all maps $f \in \mathcal{F}_{\hat{S}}$,*

$$\mathbb{E}_{x,y}[\ell(f(x), y)] \leq \frac{1}{m}\sum_{i=1}^{m}\ell(f(x_i), y_i) + 2\mathcal{R}_m(\ell\circ\mathcal{F}_{\hat{S}}) + C\sqrt{\frac{\ln(1/\delta)}{2m}}.$$

**Proof.** Define

$$\varphi(S) = \sup_{f\in\mathcal{F}_{\hat{S}}}\mathbb{E}_{x,y}[\ell(f(x), y)] - \frac{1}{m}\sum_{i=1}^{m}\ell(f(x_i), y_i).$$

To apply McDiarmid's inequality to $\varphi(S)$, we compute an upper bound on $|\varphi(S) - \varphi(S')|$ where $S = ((x_i, y_i))_{i=1}^m$ and $S' = ((x_i', y_i'))_{i=1}^m$ are two labeled datasets differing by exactly one point of an arbitrary index $i_0$; i.e., $S_i = S_i'$ for all $i \neq i_0$ and $S_{i_0} \neq S_{i_0}'$. Then,

$$\varphi(S') - \varphi(S) \leq \sup_{f\in\mathcal{F}_{\hat{S}}}\frac{\ell(f(x_{i_0}), y_{i_0}) - \ell(f(x_{i_0}'), y_{i_0}')}{m} \leq \frac{C}{m},$$

where we used that fact that $\hat{S}$ and $S$ are independent. Similarly, $\varphi(S) - \varphi(S') \leq \frac{C}{m}$. Notice that these steps fail if $\mathcal{F}_{\hat{S}}$ depends on $S$.

Thus, by McDiarmid's inequality, for any $\delta > 0$, with probability at least $1-\delta$,

$$\varphi(S) \leq \mathbb{E}_S[\varphi(S)] + C\sqrt{\frac{\ln(1/\delta)}{2m}}.$$

Moreover,

$$\begin{aligned}\mathbb{E}_S[\varphi(S)] &= \mathbb{E}_S\left[\sup_{f\in\mathcal{F}_{\hat{S}}}\mathbb{E}_{S'}\left[\frac{1}{m}\sum_{i=1}^{m}\ell(f(x_i'), y_i')\right] - \frac{1}{m}\sum_{i=1}^{m}\ell(f(x_i), y_i)\right] \\ &\leq \mathbb{E}_{S,S'}\left[\sup_{f\in\mathcal{F}_{\hat{S}}}\frac{1}{m}\sum_{i=1}^{m}(\ell(f(x_i'), y_i') - \ell(f(x_i), y_i))\right] \\ &\leq \mathbb{E}_{\xi,S,S'}\left[\sup_{f\in\mathcal{F}_{\hat{S}}}\frac{1}{m}\sum_{i=1}^{m}\xi_i(\ell(f(x_i'), y_i') - \ell(f(x_i), y_i))\right] \\ &\leq 2\mathbb{E}_{\xi,S}\left[\sup_{f\in\mathcal{F}_{\hat{S}}}\frac{1}{m}\sum_{i=1}^{m}\xi_i\ell(f(x_i), y_i))\right] = 2\mathcal{R}_m(\ell\circ\mathcal{F}_{\hat{S}})\end{aligned}$$

where the first line follows the definitions of each term, the second line uses the Jensen's inequality and the convexity of the supremum function, and the third line follows that for each $\xi_i \in \{-1, +1\}$, the distribution of each term $\xi_i(\ell(f(x_i'), y_i') - \ell(f(x_i), y_i))$ is the distribution of $(\ell(f(x_i'), y_i') - \ell(f(x_i), y_i))$ since $\bar{S}$ and $\bar{S}'$ are drawn i.i.d. with the same distribution. The fourth line uses the subadditivity of the supremum function. □

Whereas Lemma 2 is applicable for a general class of loss functions, we now recall a concrete version of Lemma 2 for binary classification. We can write the standard 0–1 loss (i.e., classification error) for binary classification by

$$\begin{aligned}\ell_{01}(f(x), y) &= \mathbb{1}\{y=1\}\mathbb{1}\{f(x) \leq 1/2\} + \mathbb{1}\{y=-1\}\mathbb{1}\{f(x) \geq 1/2\} \\ &= \mathbb{1}\{y(2f(x)-1) \leq 0\}.\end{aligned}\tag{17}$$

For any $y \in \{-1, +1\}$, we define the margin loss $\ell_\rho(f(x), y) = \bar{\ell}_\rho(y(2f(x)-1))$ as follows:

$$\bar{\ell}_\rho(a) = \begin{cases}0 & \text{if } a \geq \rho \\ 1 - a/\rho & \text{if } 0 \leq a \leq \rho \\ 1 & \text{if } a \leq 0.\end{cases}$$

Note that for any $\rho > 0$, the margin loss $\ell_\rho(yf(x))$ is an upper bound on the 0–1 loss: i.e., $\ell_\rho(f(x), y) \geq \ell_{01}(f(x), y)$. We can instantiate Lemma 2 for this concrete choice of the loss functions by using the arguments in Mohri et al. (2012):

**Lemma 3.** *Let $\mathcal{F}_{\hat{S}}$ be a set of maps $x \mapsto f(x)$ that depends on an unlabeled dataset $\hat{S}$. Fix $\rho > 0$. Then, for any $\delta > 0$, with probability at least $1-\delta$ over an i.i.d. draw of m samples $((x_i, y_i))_{i=1}^m$, each of the following holds: for all maps $f \in \mathcal{F}_{\hat{S}}$,*

$$\mathbb{E}_{x,y}[\ell_{01}(f(x), y)] \leq \frac{1}{m}\sum_{i=1}^m \ell_\rho(f(x_i), y_i) + 2\rho^{-1}\mathcal{R}_m(\mathcal{F}_{\hat{S}}) + \sqrt{\frac{\ln(1/\delta)}{2m}}, \text{ and}$$

$$\mathbb{E}_{x,y}[\ell_{01}(f(x), y)] \leq \frac{1}{m}\sum_{i=1}^m \ell_\rho(f(x_i), y_i) + 2\rho^{-1}\hat{\mathcal{R}}_m(\mathcal{F}_{\hat{S}}) + 3\sqrt{\frac{\ln(2/\delta)}{2m}}.$$

**Proof.** By combining Lemma 2 and the fact that $\ell_\rho(f(x), y) \geq \ell_{01}(f(x), y)$ and $\ell_\rho(f(x), y) \leq 1$, we have that for any $\delta > 0$, with probability at least $1-\delta$,

$$\mathbb{E}_{x,y}[\ell_{01}(f(x), y)] \leq \mathbb{E}_{x,y}[\ell_\rho(f(x), y)] \leq \frac{1}{m}\sum_{i=1}^m \ell_\rho(f(x_i), y_i) + 2\mathcal{R}_m(\ell_\rho \circ \mathcal{F}_{\hat{S}}) + \sqrt{\frac{\ln(1/\delta)}{2m}}.$$

Using Lemma 5 in Appendix A and the fact that $\ell_\rho(f(x), y)$ is $1/\rho$-Lipschitz,

$$\mathbb{E}_{x,y}[\ell_{01}(f(x), y)] \leq \frac{1}{m}\sum_{i=1}^m \ell_\rho(f(x_i), y_i) + 2\rho^{-1}\mathcal{R}_m(\mathcal{F}_{\hat{S}}) + \sqrt{\frac{\ln(1/\delta)}{2m}}.$$

This proves the first statement. For the second statement, we replace $\delta$ by $\delta/2$, use Lemma 4 in Appendix A, and take a union bound, yielding that with probability at least $1-\delta/2-\delta/2 = 1-\delta$,

$$\begin{aligned}\mathbb{E}_{x,y}[\ell_{01}(f(x), y)] &\leq \frac{1}{m}\sum_{i=1}^m \ell_\rho(f(x_i), y_i) + 2\rho^{-1}\mathcal{R}_m(\mathcal{F}_{\hat{S}}) + \sqrt{\frac{\ln(2/\delta)}{2m}} \\ &\leq \frac{1}{m}\sum_{i=1}^m \ell_\rho(f(x_i), y_i) + 2\rho^{-1}\hat{\mathcal{R}}_m(\mathcal{F}_{\hat{S}}) + 3\sqrt{\frac{\ln(2/\delta)}{2m}}\end{aligned}$$

This proves the second statement. □

Given these lemmas, we are ready to investigate how the overfitting *at labeled points* can be mitigated by using the hypothesis space $\mathcal{F}_{\hat{S},\tau}$ with the regularization on the derivatives of all orders *at unlabeled points*:

$$\mathcal{F}_{\hat{S},\tau} = \{x \mapsto f(x) : \forall \bar{S} \subseteq \hat{S}, \forall (u, k) \in \bar{S} \times \mathbb{N}^+, \|\operatorname{vec}[\partial^k f(u)]\|_2 \leq \tau_{\bar{S}}\},$$

where $\tau = \{\tau_{\bar{S}} \in \mathbb{R} : \bar{S} \subseteq \hat{S}\}$ and $\tau_{\bar{S}}$ measures the norm of the derivatives of all orders at unlabeled points. For each $\bar{S} \subseteq \hat{S}$, we write $\bar{S} = (u_1^{\bar{S}}, \ldots, u_{|\bar{S}|}^{\bar{S}})$. We define $R_{S,\bar{S}} = \max_{x\in S}\min_{u\in\bar{S}} \|x-u\|_2$, $\mathcal{S}_{S,\hat{S}} = \{\bar{S} \subseteq \hat{S} : R_{S,\bar{S}} < 1\}$, and

$$\mathcal{I}_t^{S,\bar{S}} = \left\{ i \in [m] : t = \operatorname*{argmin}_{t\in\{1,\ldots,|\bar{S}|\}} \|x_i - u_t^{\bar{S}}\|_2 \right\}.$$

The following theorem shows that regularizing $\tau_{\bar{S}}$ – the norm of the derivatives of all orders at unlabeled points – can help reducing overfitting at labeled points:

**Theorem 2.** *Fix $\mathcal{F}_{\hat{S},\tau}$ and $\rho > 0$. Then, for any $\delta > 0$, with probability at least $1-\delta$ over an i.i.d. draw of m samples $((x_i, y_i))_{i=1}^m$, each of the following holds: for all maps $f \in \mathcal{F}_{\hat{S},\tau}$,*

$$\begin{aligned}&\mathbb{E}_{x,y}[\ell_{01}(f(x), y)] \\ &\leq \frac{1}{m}\sum_{i=1}^m \ell_\rho(f(x_i), y_i) + 2\rho^{-1}\mathbb{E}_S \\ &\times \left[ \min_{\bar{S}\in\mathcal{S}_{S,\hat{S}}} \left(1 + \frac{\tau_{\bar{S}} R_{S,\bar{S}}}{1 - R_{S,\bar{S}}}\right) \frac{\sum_{t=1}^{|\bar{S}|}\sqrt{|\mathcal{I}_t^{S,\bar{S}}|}}{m} \right] + \sqrt{\frac{\ln(1/\delta)}{2m}},\end{aligned}$$

*and*

$$\begin{aligned}&\mathbb{E}_{x,y}[\ell_{01}(f(x), y)] \\ &\leq \frac{1}{m}\sum_{i=1}^m \ell_\rho(f(x_i), y_i) + 2\rho^{-1} \min_{\bar{S}\in\mathcal{S}_{S,\hat{S}}} \left(1 + \frac{\tau_{\bar{S}} R_{S,\bar{S}}}{1 - R_{S,\bar{S}}}\right) \frac{\sum_{t=1}^{|\bar{S}|}\sqrt{|\mathcal{I}_t^{S,\bar{S}}|}}{m} \\ &+ 3\sqrt{\frac{\ln(2/\delta)}{2m}}.\end{aligned}$$

**Proof.** Let $\bar{S} \subseteq \hat{S}$ be arbitrary such that $R_{S,\bar{S}} < 1$. Using Lemma 1 and the Cauchy–Schwarz inequality,

$$\sum_{k=1}^K \operatorname{vec}[\partial^k f(u)]^\top (x-u)^{\otimes k} \leq \sum_{k=1}^K \|\operatorname{vec}[\partial^k f(u)]\|_2 \|x-u\|_2^k.$$

Thus, for any $(f, u) \in \mathcal{F}_{\hat{S},\tau} \times \bar{S}$,

$$\lim_{K\to\infty} \sum_{k=1}^K \|\operatorname{vec}[\partial^k f(u)]\|_2 \|x-u\|_2^k \leq \tau_{\bar{S}} \lim_{K\to\infty} \sum_{k=1}^K \|x-u\|_2^k.$$

Since $\lim_{K\to\infty}\sum_{k=1}^K \|x-u\|_2^k$ is a geometric series, if $\|x-u\|_2 \leq R < 1$, then

$$\lim_{K\to\infty} \sum_{k=1}^K \|x-u\|_2^k \leq \frac{1}{1-R} - 1.$$

Therefore, combining above inequalities, for any $(f, u) \in \mathcal{F}_{\hat{S},\tau} \times \bar{S}$, if $\|x-u\|_2 \leq R < 1$,

$$\sum_{k=1}^\infty \operatorname{vec}[\partial^k f(u)]^\top (x-u)^{\otimes k} \leq \frac{\tau_{\bar{S}}}{1-R} - 1. \tag{18}$$

Let $\hat{t}(x) = \operatorname{argmin}_{t\in\{1,\ldots,|\bar{S}|\}} \|x - u_t^{\bar{S}}\|_2$. Then,

$$\hat{\mathcal{R}}_m(\mathcal{F}_{\hat{S},\tau}) = \mathbb{E}_\xi \sup_{f\in\mathcal{F}_{\hat{S},\tau}} \frac{1}{m}\sum_{i=1}^m \xi_i f(x_i) = \mathbb{E}_\xi \sup_{f\in\mathcal{F}_{\hat{S},\tau}} \frac{1}{m}\sum_{i=1}^m \xi_i f(u_{\hat{t}(x_i)}^{\bar{S}} + (x_i - u_{\hat{t}(x_i)}^{\bar{S}})).$$

Thus, Eq. (18) implies that

$$\begin{aligned}\hat{\mathcal{R}}_m(\mathcal{F}_{\hat{S},\tau}) &= \mathbb{E}_\xi \sup_{f\in\mathcal{F}_{\hat{S},\tau}} \frac{1}{m}\sum_{i=1}^m \xi_i f(u_{\hat{t}(x_i)}^{\bar{S}} + (x - u_{\hat{t}(x_i)}^{\bar{S}})) \\ &= \mathbb{E}_\xi \sup_{f\in\mathcal{F}_{\hat{S},\tau}} \frac{1}{m}\sum_{i=1}^m \xi_i \left(f(u_{\hat{t}(x_i)}^{\bar{S}}) + \sum_{k=1}^\infty \operatorname{vec}[\partial^k f(u_{\hat{t}(x_i)}^{\bar{S}})]^\top (x_i - u_{\hat{t}(x_i)}^{\bar{S}})^{\otimes k}\right)\end{aligned}$$

where the series converges based on Eq. (18) since $\min_{u\in\bar{S}} \|x-u\|_2 < 1$ for all $x \in S_x$. Since $\mathcal{I}_t^{S,\bar{S}} = \{i \in [m] : t = \operatorname{argmin}_{t\in\{1,\ldots,|\bar{S}|\}} \|x_i - u_t^{\bar{S}}\|_2\} = \{i \in [m] : t = \hat{t}(x_i)\}$, we can further rewrite

$$\begin{aligned}&\hat{\mathcal{R}}_m(\mathcal{F}_{\hat{S},\tau}) \\ &= \mathbb{E}_\xi \sup_{f\in\mathcal{F}_{\hat{S},\tau}} \frac{1}{m}\sum_{i=1}^m \xi_i \left(f(u_{\hat{t}(x_i)}^{\bar{S}}) + \sum_{k=1}^\infty \operatorname{vec}[\partial^k f(u_{\hat{t}(x_i)}^{\bar{S}})]^\top (x_i - u_{\hat{t}(x_i)}^{\bar{S}})^{\otimes k}\right) \\ &= \mathbb{E}_\xi \sup_{f\in\mathcal{F}_{\hat{S},\tau}} \frac{1}{m}\sum_{t=1}^{|\bar{S}|}\sum_{i\in\mathcal{I}_t^{S,\bar{S}}} \xi_i \left(f(u_{\hat{t}(x_i)}^{\bar{S}}) + \sum_{k=1}^\infty \operatorname{vec}[\partial^k f(u_{\hat{t}(x_i)}^{\bar{S}})]^\top (x_i - u_{\hat{t}(x_i)}^{\bar{S}})^{\otimes k}\right)\end{aligned}$$

$$= \mathbb{E}_\xi \sup_{f\in\mathcal{F}_{\hat{S},\tau}} \frac{1}{m}\sum_{t=1}^{|\bar{S}|}\sum_{i\in\mathcal{I}_t^{S,\bar{S}}} \xi_i \left(f(u_t^{\bar{S}}) + \sum_{k=1}^{\infty} \text{vec}[\partial^k f(u_t^{\bar{S}})]^\top (x_i - u_t^{\bar{S}})^{\otimes k}\right)$$

$$= \frac{1}{m}\mathbb{E}_\xi \sup_{f\in\mathcal{F}_{\hat{S},\tau}} \left[\sum_{t=1}^{|\bar{S}|}\sum_{i\in\mathcal{I}_t^{S,\bar{S}}} \xi_i f(u_t^{\bar{S}}) + \sum_{t=1}^{|\bar{S}|}\sum_{i\in\mathcal{I}_t^{S,\bar{S}}} \xi_i \sum_{k=1}^{\infty} \text{vec}[\partial^k f(u_t^{\bar{S}})]^\top (x_i - u_t^{\bar{S}})^{\otimes k}\right]$$

$$\leq \frac{1}{m}\mathbb{E}_\xi \left[\sum_{t=1}^{|\bar{S}|}\sup_{f\in\mathcal{F}_{\hat{S},\tau}} |f(u_t^{\bar{S}})| \left|\sum_{i\in\mathcal{I}_t^{S,\bar{S}}} \xi_i\right| + \sum_{t=1}^{|\bar{S}|}\sum_{k=1}^{\infty}\sup_{f\in\mathcal{F}_{\hat{S},\tau}} \|\text{vec}[\partial^k f(u_t^{\bar{S}})]\|_2 \left\|\sum_{i\in\mathcal{I}_t^{S,\bar{S}}} \xi_i (x_i - u_t)^{\otimes k}\right\|_2\right]$$

Using the definition of the function class $\mathcal{F}_{\hat{S},\tau}$ and the fact that $|f(u_t^{\bar{S}})| \leq 1$, we have

$$\hat{\mathcal{R}}_m(\mathcal{F}_{\hat{S},\tau})$$

$$\leq \frac{1}{m}\mathbb{E}_\xi \left[\sum_{t=1}^{|\bar{S}|}\sup_{f\in\mathcal{F}_{\hat{S},\tau}} |f(u_t^{\bar{S}})| \left|\sum_{i\in\mathcal{I}_t^{S,\bar{S}}} \xi_i\right| + \sum_{t=1}^{|\bar{S}|}\sum_{k=1}^{\infty}\sup_{f\in\mathcal{F}_{\hat{S},\tau}} \|\text{vec}[\partial^k f(u_t^{\bar{S}})]\|_2 \left\|\sum_{i\in\mathcal{I}_t^{S,\bar{S}}} \xi_i (x_i - u_t^{\bar{S}})^{\otimes k}\right\|_2\right]$$

$$\leq \frac{1}{m}\mathbb{E}_\xi \left[\sum_{t=1}^{|\bar{S}|}\left|\sum_{i\in\mathcal{I}_t^{S,\bar{S}}} \xi_i\right| + \tau_{\bar{S}}\sum_{t=1}^{|\bar{S}|}\sum_{k=1}^{\infty}\left\|\sum_{i\in\mathcal{I}_t^{S,\bar{S}}} \xi_i (x_i - u_t^{\bar{S}})^{\otimes k}\right\|_2\right]$$

$$= \frac{1}{m}\sum_{t=1}^{|\bar{S}|}\mathbb{E}_\xi \sqrt{\left(\sum_{i\in\mathcal{I}_t^{S,\bar{S}}} \xi_i\right)^2} + \frac{\tau_{\bar{S}}}{m}\sum_{t=1}^{|\bar{S}|}\sum_{k=1}^{\infty}\mathbb{E}_\xi\left\|\sum_{i\in\mathcal{I}_t^{S,\bar{S}}} \xi_i (x_i - u_t^{\bar{S}})^{\otimes k}\right\|_2 \tag{19}$$

where the last line follows the linearity of the expectation. By using Jensen's inequality for the concave function,

$$\mathbb{E}_\xi \sqrt{\left(\sum_{i\in\mathcal{I}_t^{S,\bar{S}}} \xi_i\right)^2} \leq \sqrt{\mathbb{E}_\xi\left(\sum_{i\in\mathcal{I}_t^{S,\bar{S}}} \xi_i\right)^2} = \sqrt{\mathbb{E}_\xi\sum_{i\in\mathcal{I}_t^{S,\bar{S}}}\sum_{j\in\mathcal{I}_t^{S,\bar{S}}} \xi_i\xi_j}$$
$$= \sqrt{\sum_{i\in\mathcal{I}_t^{S,\bar{S}}}\sum_{j\in\mathcal{I}_t^{S,\bar{S}}} \mathbb{E}_\xi[\xi_i\xi_j]} = \sqrt{\sum_{i\in\mathcal{I}_t^{S,\bar{S}}} \mathbb{E}_\xi[\xi_i^2]}$$

where the last line follows the fact that the Rademacher variables $\xi_1, \ldots, \xi_m$ are independent. Since $\mathbb{E}_\xi[\xi_i^2] = 1$,

$$\mathbb{E}_\xi \sqrt{\left(\sum_{i\in\mathcal{I}_t^{S,\bar{S}}} \xi_i\right)^2} \leq \sqrt{|\mathcal{I}_t^{S,\bar{S}}|} \tag{20}$$

Similarly, by using Jensen's inequality for the concave function,

$$\mathbb{E}_\xi\left\|\sum_{i\in\mathcal{I}_t^{S,\bar{S}}} \xi_i (x_i - u_t^{\bar{S}})^{\otimes k}\right\|_2 = \mathbb{E}_\xi\sqrt{\sum_{i\in\mathcal{I}_t^{S,\bar{S}}}\sum_{j\in\mathcal{I}_t^{S,\bar{S}}} \xi_i\xi_j ((x_i - u_t^{\bar{S}})^{\otimes k})^\top (x_j - u_t^{\bar{S}})^{\otimes k}}$$
$$\leq \sqrt{\sum_{i\in\mathcal{I}_t^{S,\bar{S}}}\sum_{j\in\mathcal{I}_t^{S,\bar{S}}} \mathbb{E}_\xi[\xi_i\xi_j]((x_i - u_t^{\bar{S}})^{\otimes k})^\top (x_j - u_t^{\bar{S}})^{\otimes k}}$$
$$\leq \sqrt{\sum_{i\in\mathcal{I}_t^{S,\bar{S}}} \mathbb{E}_\xi[\xi_i^2]\|(x_i - u_t^{\bar{S}})^{\otimes k}\|_2^2}$$

where the last line follows the fact that Rademacher variables $\xi_1, \ldots, \xi_m$ are independent. Since $\mathbb{E}_\xi[\xi_i^2] = 1$, using Lemma 1,

$$\mathbb{E}_\xi\left\|\sum_{i\in\mathcal{I}_t^{S,\bar{S}}} \xi_i (x_i - u_t^{\bar{S}})^{\otimes k}\right\|_2 \leq \sqrt{\sum_{i\in\mathcal{I}_t^{S,\bar{S}}} \|(x_i - u_t^{\bar{S}})^{\otimes k}\|_2^2}$$
$$= \sqrt{\sum_{i\in\mathcal{I}_t^{S,\bar{S}}} (\|x_i - u_t^{\bar{S}}\|_2)^{2k}}$$
$$\leq R_{S,\bar{S}}^k\sqrt{|\mathcal{I}_t^{S,\bar{S}}|} \tag{21}$$

Combining inequalities (19)–(21) yields

$$m\hat{\mathcal{R}}_m(\mathcal{F}_{\hat{S},\tau}) \leq \sum_{t=1}^{|\bar{S}|}\sqrt{|\mathcal{I}_t^{S,\bar{S}}|} + \tau_{\bar{S}}\sum_{t=1}^{|\bar{S}|}\sqrt{|\mathcal{I}_t^{S,\bar{S}}|}\sum_{k=1}^{\infty} R_{S,\bar{S}}^k.$$

Since $R_{S,\bar{S}} < 1$, the geometric series converges as

$$m\hat{\mathcal{R}}_m(\mathcal{F}_{\hat{S},\tau}) \leq \sum_{t=1}^{|\bar{S}|}\sqrt{|\mathcal{I}_t^{S,\bar{S}}|} + \frac{\tau_{\bar{S}}R_{S,\bar{S}}}{1-R_{S,\bar{S}}}\sum_{t=1}^{|\bar{S}|}\sqrt{|\mathcal{I}_t^{S,\bar{S}}|}$$
$$= \left(1 + \frac{\tau_{\bar{S}}R_{S,\bar{S}}}{1-R_{S,\bar{S}}}\right)\sum_{t=1}^{|\bar{S}|}\sqrt{|\mathcal{I}_t^{S,\bar{S}}|}.$$

Therefore,

$$\hat{\mathcal{R}}_m(\mathcal{F}_{\hat{S},\tau}) \leq \left(1 + \frac{\tau_{\bar{S}}R_{S,\bar{S}}}{1-R_{S,\bar{S}}}\right)\frac{\sum_{t=1}^{|\bar{S}|}\sqrt{|\mathcal{I}_t^{S,\bar{S}}|}}{m}.$$

Since $\bar{S} \subseteq \hat{S}$ was arbitrary such that $R_{S,\bar{S}} < 1$, this inequality holds for any $\bar{S}$ such that $R_{S,\bar{S}} < 1$, yielding that for any $S = ((x_1, y_1), \ldots, (x_m, y_m))$,

$$\hat{\mathcal{R}}_m(\mathcal{F}_{\hat{S},\tau}) = \mathbb{E}_\xi \sup_{f\in\mathcal{F}_{\hat{S},\tau}}\frac{1}{m}\sum_{i=1}^m \xi_i f(x_i) \leq \min_{\bar{S}\in\mathcal{S}_{S,\hat{S}}}\left(1 + \frac{\tau_{\bar{S}}R_{S,\bar{S}}}{1-R_{S,\bar{S}}}\right) \times \frac{\sum_{t=1}^{|\bar{S}|}\sqrt{|\mathcal{I}_t^{S,\bar{S}}|}}{m}. \tag{22}$$

By combining Lemma 3 and inequality (22) and taking expectation over $S$, we obtain the statement of this theorem. □

Since $\min_{\bar{S}\in\mathcal{S}_{S,\hat{S}}} g(\bar{S}) \leq g(\bar{S}')$ for any $\bar{S}' \in \mathcal{S}_{S,\hat{S}}$ and any function $g$, Theorem 2 implies that with probability at least $1-\delta$, for any $\bar{S} \in \mathcal{S}_{S,\hat{S}}$,

$$\mathbb{E}_{x,y}[\ell_{01}(f(x), y)]$$
$$\leq \frac{1}{m}\sum_{i=1}^m \ell_\rho(f(x_i), y_i) + 2\rho^{-1}\left(1 + \frac{\tau_{\bar{S}}R_{S,\bar{S}}}{1-R_{S,\bar{S}}}\right)\frac{\sum_{t=1}^{|\bar{S}|}\sqrt{|\mathcal{I}_t^{S,\bar{S}}|}}{m} + 3\sqrt{\frac{\ln(2/\delta)}{2m}}.$$

To further understand Theorem 2, let us consider a simple case where the input space is normalized such that the maximum distance between some unlabeled point $u_c \in \hat{S}$ and the labeled points $S$ is bounded as $\max_{x\in S}\|x - u_c\|_2 < 1/2$. Then, by setting

$\bar{S} = \{u_c\}$, Theorem 2 implies that with probability at least $1-\delta$,

$$\mathbb{E}_{x,y}[\ell_{01}(f(x),y)] \leq \frac{1}{m}\sum_{i=1}^{m}\ell_\rho(f(x_i),y_i) + \frac{2\rho^{-1}(1+\tau_{\bar{S}})}{\sqrt{m}} + 3\sqrt{\frac{\ln(2/\delta)}{2m}}, \tag{23}$$

since $R_{S,\bar{S}} = \frac{1}{2}$, $|\bar{S}| = 1$, and $|\mathcal{I}_t^{S,\bar{S}}| = m$ for this particular choice of $\bar{S} = \{u_c\}$. In Eq. (23), we can clearly see that the degree of possible overfitting at labeled points is in the order of $\frac{1+\tau_{\bar{S}}}{\sqrt{m}}$, which can be reduced by regularizing $\tau_{\bar{S}}$ — the norm of the derivatives of all orders at unlabeled points.

In Eq. (23) (and Theorem 2), there is a tradeoff between reducing $\tau_{\bar{S}}$ and minimizing the classification loss $\frac{1}{m}\sum_{i=1}^{m}\ell_\rho(f(x_i),y_i)$. In an extreme case, if we minimize $\tau_{\bar{S}}$ to zero, then we can minimize the order of the overfitting from $\frac{1+\tau_{\bar{S}}}{\sqrt{m}}$ to $\frac{1}{\sqrt{m}}$ but the model $f$ becomes a constant function which cannot minimize the classification loss $\frac{1}{m}\sum_{i=1}^{m}\ell_\rho(f(x_i),y_i)$ in typical practical applications.

Whereas this tradeoff is natural in the regime of large $\tau_{\bar{S}}$, it is not desirable to require $f$ to be a constant function in order to obtain a global minimum value of the degree of the overfitting in the regime of small $\tau_{\bar{S}}$. Accordingly, we now prove an additional theorem to avoid this tradeoff in the regime of small $\tau_{\bar{S}}$. More concretely, Theorem 3 shows that we can reduce the order of the overfitting from $\frac{1+\tau_{\bar{S}}}{\sqrt{m}}$ to $\frac{1}{\sqrt{m}}$ without requiring $\tau_{\bar{S}} = 0$ in the regime where $\tau_{\bar{S}}$ is smaller than some confidence value at unlabeled points as $\tau_{\bar{S}} < \mathcal{C}_{\bar{S}}\frac{(1-R_{S,\bar{S}})}{R_{S,\bar{S}}}$, where $\mathcal{C}_{\bar{S}}$ measures the confidence values at unlabeled points. Thus, Theorem 3 also shows the benefit of increasing confidence values at unlabeled points, which is consistent with our observation in Fig. 3.

**Theorem 3.** *Fix $\mathcal{F}_{\hat{S},\tau}$. Define $\mathcal{C}_{\bar{S}} = \inf_{f\in\mathcal{F}_{\hat{S},\tau}} \min_{u\in\bar{S}} |f(u)-1/2|$ and $\mathcal{S}^*_{S,\hat{S}} = \{\bar{S}\subseteq\hat{S} : R_{S,\bar{S}} < 1, \tau_{\bar{S}} < \frac{\mathcal{C}_{\bar{S}}(1-R_{S,\bar{S}})}{R_{S,\bar{S}}}\}$. Then, for any $\delta > 0$, with probability at least $1-\delta$ over an i.i.d. draw of $m$ samples $((x_i,y_i))_{i=1}^m$, each of the following holds: for all maps $f\in\mathcal{F}_{\hat{S},\tau}$,*

$$\mathbb{E}_{x,y}[\ell_{01}(f(x),y)] \leq \frac{1}{m}\sum_{i=1}^{m}\ell_{01}(f(x),y) + \frac{\mathbb{E}_S\left[\min_{\bar{S}\in\mathcal{S}^*_{S,\hat{S}}}\sum_{t=1}^{|\bar{S}|}\sqrt{|\mathcal{I}_t^{S,\bar{S}}|}\right]}{m} + \sqrt{\frac{\ln(2/\delta)}{2m}},$$

$$\mathbb{E}_{x,y}[\ell_{01}(f(x),y)] \leq \frac{1}{m}\sum_{i=1}^{m}\ell_{01}(f(x),y) + \frac{\min_{\bar{S}\in\mathcal{S}^*_{S,\hat{S}}}\sum_{t=1}^{|\bar{S}|}\sqrt{|\mathcal{I}_t^{S,\bar{S}}|}}{m} + 3\sqrt{\frac{\ln(1/\delta)}{2m}}.$$

**Proof.** Without the loss of generality, let us write $y\in\{-1,+1\}$ (if the original label $y$ is in $\{0,1\}$, then we can define a bijection to map $\{0,1\}$ to $\{-1,+1\}$). Define

$$\varsigma(f(x)) = \begin{cases} +1 & \text{if } f(x)-\frac{1}{2} > 0 \\ -1 & \text{if } f(x)-\frac{1}{2} \leq 0 \end{cases}$$

Then, we can write the 0–1 loss of the classification as

$$\ell_{01}(f(x),y) = \mathbb{1}\{\varsigma(f(x)) \neq y\} = \frac{1-y\varsigma(f(x))}{2} \tag{24}$$

By using Lemma 2 with the 0–1 loss, we have that for any $\delta > 0$, with probability at least $1-\delta$, for all $f\in\mathcal{F}_{\hat{S},\tau}$,

$$\mathbb{E}_{x,y}[\ell_{01}(f(x),y)] \leq \frac{1}{m}\sum_{i=1}^{m}\ell_{01}(f(x),y) + 2\mathcal{R}_m(\ell_{01}\circ\mathcal{F}_{\hat{S},\tau}) + \sqrt{\frac{\ln(2/\delta)}{2m}}. \tag{25}$$

By using Lemma 4 and taking a union bound, we have that with probability at least $1-\delta/2-\delta/2 = 1-\delta$, for all $f\in\mathcal{F}_{\hat{S},\tau}$,

$$\mathbb{E}_{x,y}[\ell_{01}(f(x),y)] \leq \frac{1}{m}\sum_{i=1}^{m}\ell_{01}(f(x),y) + 2\hat{\mathcal{R}}_m(\ell_{01}\circ\mathcal{F}_{\hat{S},\tau}) + 3\sqrt{\frac{\ln(1/\delta)}{2m}}. \tag{26}$$

Using Eq. (24),

$$\begin{aligned}
\hat{\mathcal{R}}_m(\ell_{01}\circ\mathcal{F}_{\bar{S}}) &= \mathbb{E}_\xi \sup_{f\in\mathcal{F}_{\hat{S},\tau}} \frac{1}{m}\sum_{i=1}^{m}\xi_i\ell_{01}(f(x_i),y_i) \\
&= \mathbb{E}_\xi \sup_{f\in\mathcal{F}_{\hat{S},\tau}} \frac{1}{m}\sum_{i=1}^{m}\xi_i\frac{1-y_i\varsigma(f(x_i))}{2} \\
&= \mathbb{E}_\xi \sup_{f\in\mathcal{F}_{\hat{S},\tau}} \frac{1}{m}\sum_{i=1}^{m}\frac{-\xi_i y_i\varsigma(f(x_i))}{2} \\
&= \frac{1}{2m}\mathbb{E}_\xi \sup_{f\in\mathcal{F}_{\hat{S},\tau}} \sum_{i=1}^{m}\xi_i\varsigma(f(x_i)).
\end{aligned}$$

Let $\bar{S}\subseteq\hat{S}$ be arbitrary such that $R_{S,\bar{S}} < 1$ and $\tau_{\bar{S}} < \frac{\mathcal{C}_{\bar{S}}(1-R_{S,\bar{S}})}{R_{S,\bar{S}}}$. Let $\hat{t}(x) = \operatorname{argmin}_{t\in\{1,\ldots|\bar{S}|\}}\|x-u_t^{\bar{S}}\|_2$. Then, for any $f\in\mathcal{F}_{\hat{S},\tau}$, the proof of Theorem 2 shows that we can write

$$f(x_i) = f(u^{\bar{S}}_{\hat{t}(x_i)}) + \sum_{k=1}^{\infty}\operatorname{vec}[\partial^k f(u^{\bar{S}}_{\hat{t}(x_i)})]^\top(x_i-u^{\bar{S}}_{\hat{t}(x_i)})^{\otimes k}.$$

Since $\mathcal{I}_t^{S,\bar{S}} = \{i\in[m] : t = \operatorname{argmin}_{t\in\{1,\ldots,|\bar{S}|\}}\|x_i-u_t^{\bar{S}}\|_2\} = \{i\in[m] : t = \hat{t}(x_i)\}$, we can rewrite

$$\begin{aligned}
\hat{\mathcal{R}}_m(\ell_{01}\circ\mathcal{F}_{\bar{S}}) &= \frac{1}{2m}\mathbb{E}_\xi \sup_{f\in\mathcal{F}_{\hat{S},\tau}} \sum_{i=1}^{m}\xi_i\varsigma \\
&\quad\times\left(f(u^{\bar{S}}_{\hat{t}(x_i)}) + \sum_{k=1}^{\infty}\operatorname{vec}[\partial^k f(u^{\bar{S}}_{\hat{t}(x_i)})]^\top(x_i-u^{\bar{S}}_{\hat{t}(x_i)})^{\otimes k}\right) \\
&= \frac{1}{2m}\mathbb{E}_\xi \sup_{f\in\mathcal{F}_{\hat{S},\tau}} \sum_{t=1}^{|\bar{S}|}\sum_{i\in\mathcal{I}_t^{S,\bar{S}}}\xi_i\varsigma \\
&\quad\times\left(f(u_t^{\bar{S}}) + \sum_{k=1}^{\infty}\operatorname{vec}[\partial^k f(u_t^{\bar{S}})]^\top(x_i-u_t^{\bar{S}})^{\otimes k}\right)
\end{aligned}$$

Using Lemma 1,

$$\begin{aligned}
\left|\sum_{k=1}^{\infty}\operatorname{vec}[\partial^k f(u_t^{\bar{S}})]^\top(x_i-u_t^{\bar{S}})^{\otimes k}\right| &\leq \sum_{k=1}^{\infty}\|\operatorname{vec}[\partial^k f(u_t^{\bar{S}})]\|_2\|(x_i-u_t^{\bar{S}})^{\otimes k}\|_2 \\
&\leq \tau_{\bar{S}}\frac{R_{S,\bar{S}}}{1-R_{S,\bar{S}}} \\
&< \mathcal{C}_{\bar{S}}
\end{aligned}$$

where the last line follows the condition on $\bar{S}$ that $\tau_{\bar{S}} < \frac{\mathcal{C}_{\bar{S}}(1-R_{S,\bar{S}})}{R_{S,\bar{S}}}$ (since $\tau_{\bar{S}}$ is defined independently of $S$, this condition does not necessarily hold for some $\bar{S} \subseteq \hat{S}$. If this does not hold for all $\bar{S} \subseteq \hat{S}$, then the statement holds vacuously with the convention that $\min_{a \in \emptyset} \Psi(a) = \infty$ for any function $\Psi$). Since $|f(u_t^{\bar{S}}) - 1/2| \geq \mathcal{C}_{\bar{S}}$ and $|\sum_{k=1}^{\infty} \mathrm{vec}[\partial^k f(u_t^{\bar{S}})]^\top (x_i - u_t^{\bar{S}})^{\otimes k}| < \mathcal{C}_{\bar{S}}$, we have that

$$\begin{aligned}
\hat{\mathcal{R}}_m(\ell_{01} \circ \mathcal{F}_{\bar{S}}) &= \frac{1}{2m} \mathbb{E}_\xi \sup_{f \in \mathcal{F}_{\hat{S},\tau}} \sum_{t=1}^{|\bar{S}|} \sum_{i \in \mathcal{I}_t^{S,\bar{S}}} \xi_i \varsigma \\
&\quad \times \left( f(u_t^{\bar{S}}) + \sum_{k=1}^{\infty} \mathrm{vec}[\partial^k f(u_t^{\bar{S}})]^\top (x_i - u_t^{\bar{S}})^{\otimes k} \right) \\
&= \frac{1}{2m} \mathbb{E}_\xi \sup_{f \in \mathcal{F}_{\hat{S},\tau}} \sum_{t=1}^{|\bar{S}|} \varsigma \left( f(u_t^{\bar{S}}) \right) \sum_{i \in \mathcal{I}_t^{S,\bar{S}}} \xi_i \\
&\leq \frac{1}{2m} \mathbb{E}_\xi \sum_{t=1}^{|\bar{S}|} \sup_{f \in \mathcal{F}_{\hat{S},\tau}} \left| \varsigma \left( f(u_t^{\bar{S}}) \right) \right| \left| \sum_{i \in \mathcal{I}_t^{S,\bar{S}}} \xi_i \right| \\
&\leq \frac{1}{2m} \sum_{t=1}^{|\bar{S}|} \mathbb{E}_\xi \sqrt{\left( \sum_{i \in \mathcal{I}_t^{S,\bar{S}}} \xi_i \right)^2}
\end{aligned}$$

By using Jensen's inequality for the concave function,

$$\begin{aligned}
\mathbb{E}_\xi \sqrt{\left( \sum_{i \in \mathcal{I}_t^{S,\bar{S}}} \xi_i \right)^2} \leq \sqrt{\mathbb{E}_\xi \left( \sum_{i \in \mathcal{I}_t^{S,\bar{S}}} \xi_i \right)^2} &= \sqrt{\sum_{i \in \mathcal{I}_t^{S,\bar{S}}} \sum_{j \in \mathcal{I}_t^{S,\bar{S}}} \mathbb{E}_\xi [\xi_i \xi_j]} \\
&= \sqrt{\sum_{i \in \mathcal{I}_t^{S,\bar{S}}} \mathbb{E}_\xi [\xi_i^2]} \\
&= \sqrt{|\mathcal{I}_t^{S,\bar{S}}|}
\end{aligned}$$

Therefore,

$$\hat{\mathcal{R}}_m(\ell_{01} \circ \mathcal{F}_{\bar{S}}) \leq \frac{1}{2m} \sum_{t=1}^{|\bar{S}|} \sqrt{|\mathcal{I}_t^{S,\bar{S}}|}.$$

Since $\bar{S} \subseteq \hat{S}$ was arbitrary such that $R_{S,\bar{S}} < 1$ and $\tau_{\bar{S}} < \frac{\mathcal{C}_{\bar{S}}(1-R_{S,\bar{S}})}{R_{S,\bar{S}}}$, this inequality holds for any $\bar{S}$ such that $R_{S,\bar{S}} < 1$ and $\tau_{\bar{S}} < \frac{\mathcal{C}_{\bar{S}}(1-R_{S,\bar{S}})}{R_{S,\bar{S}}}$, yielding

$$\hat{\mathcal{R}}_m(\ell_{01} \circ \mathcal{F}_{\bar{S}}) \leq \frac{1}{2m} \min_{\bar{S} \in \mathcal{S}^*_{S,\hat{S}}} \sum_{t=1}^{|\bar{S}|} \sqrt{|\mathcal{I}_t^{S,\bar{S}}|}.$$

Taking expectation and combining with Eqs. (25)–(26) yields that for any $\delta > 0$, with probability at least $1-\delta$, each of the following holds for all $f \in \mathcal{F}_{\hat{S},\tau}$:

$$\begin{aligned}
\mathbb{E}_{x,y}[\ell_{01}(f(x), y)] &\leq \frac{1}{m} \sum_{i=1}^{m} \ell_{01}(f(x), y) \\
&\quad + \frac{\mathbb{E}_S \left[ \min_{\bar{S} \in \mathcal{S}^*_{S,\hat{S}}} \sum_{t=1}^{|\bar{S}|} \sqrt{|\mathcal{I}_t^{S,\bar{S}}|} \right]}{m} + \sqrt{\frac{\ln(2/\delta)}{2m}}. \\
\mathbb{E}_{x,y}[\ell_{01}(f(x), y)] &\leq \frac{1}{m} \sum_{i=1}^{m} \ell_{01}(f(x), y) \\
&\quad + \frac{\min_{\bar{S} \in \mathcal{S}^*_{S,\hat{S}}} \sum_{t=1}^{|\bar{S}|} \sqrt{|\mathcal{I}_t^{S,\bar{S}}|}}{m} + 3\sqrt{\frac{\ln(1/\delta)}{2m}}. \quad \square
\end{aligned}$$

Since $\min_{\bar{S} \in \mathcal{S}^*_{S,\hat{S}}} g(\bar{S}) \leq g(\bar{S}')$ for any $\bar{S}' \in \mathcal{S}^*_{S,\hat{S}}$ and any function $g$, Theorem 3 implies that with probability at least $1 - \delta$, for any $\bar{S} = \{u_c\} \in \mathcal{S}^*_{S,\hat{S}}$,

$$\mathbb{E}_{x,y}[\ell_{01}(f(x), y)] \leq \frac{1}{m} \sum_{i=1}^{m} \ell_{01}(f(x), y) + \frac{1}{\sqrt{m}} + 3\sqrt{\frac{\ln(1/\delta)}{2m}}, \quad (27)$$

as $|\bar{S}| = 1$ and $|\mathcal{I}_t^{S,\bar{S}}| = m$ for the singleton set $\bar{S} = \{u_c\}$. Therefore, if we increase the confidence at unlabeled points and regularize the norm of the derivatives of all orders at unlabeled points, we can reduce overfitting at labeled points: i.e., the classification error $\frac{1}{m} \sum_{i=1}^{m} \ell_{01}(f(x), y)$ at labeled points approaches $\mathbb{E}_{x,y}[\ell_{01}(f(x), y)]$ in the rate of $O(\sqrt{\ln(1/\delta)/m})$.

Finally, we remark that our theories are reflecting the fact that the prediction at unlabeled points could be wrong, although the confidence can be high. Indeed, in all of our experiments and theoretical analyses, we do *not* assume that the pseudo labels at unlabeled points are correct or of good quality. Instead, our proofs analyze the possible cases of the bad quality of pseudo-labels and show the following: a prediction of ICT at an unlabeled point is likely correct if it has a high confidence at the unlabeled point *and* if the unlabeled point is near a correctly classified labeled point. Intuitively, this is because when the derivatives of all orders are regularized, the prediction cannot change a lot with respect to the input. This is one of ways how labeled and unlabeled points are interacted in our new proof techniques in this paper. We expect our new proof techniques to be proven useful in future studies of semi-supervised learning.

In Theorems 2 and 3, we can see that if all unlabeled points are too far way from the labeled points, then the bounds degrade linearly in the distance between the labeled and unlabeled points. This is because the prediction could be more likely inaccurate as this distance increases. Here, by normalizing the input space, we can guarantee for this distance to be sufficiently small. As an example, the discussions after Theorem 2 provide the case where the input space is normalized such that the maximum distance between a center unlabeled point $u_c \in \hat{S}$ and labeled points $S$ is bounded as $\max_{x \in S} \|x - u_c\|_2 < 1/2$.

More importantly, the distance between the labeled and unlabeled points that is used in Theorems 2 and 3 tends to decrease as we increase only the number of *unlabeled* points. This is because the distance is defined by $R_{S,\hat{S}} = \max_{x \in S} \min_{u \in \hat{S}} \|x - u\|_2$ with the minimum over all the unlabeled points $u \in \hat{S}$, for example, with $\bar{S} = \hat{S}$. Therefore, the quality of pseudo labels at unlabeled points can improve only by increasing the number of *unlabeled* points. This theoretical prediction is consistent with our main experimental results in Section 3 and is further validated via the additional experiments in Appendix B.2.

## 5. Related work

This work builds on two threads of research: consistency-regularization for semi-supervised learning and interpolation-based regularizers.

Consistency-regularization semi-supervised learning methods (Athiwaratkun et al., 2019; Laine & Aila, 2017; Luo et al., 2018; Miyato et al., 2018; Sajjadi et al., 2016; Tarvainen & Valpola, 2017) encourage that realistic perturbations $u + \delta$ of unlabeled samples $u$ should not change the model predictions $f_\theta(u)$. Motivated by the *low-density separation assumption* (Chapelle et al., 2010), these methods push the decision boundary towards the low-density regions of the input space, achieving larger classification margins. ICT differs from these approaches in two aspects. First, ICT chooses perturbations in the direction of another randomly chosen unlabeled sample, avoiding expensive gradient computations.

When interpolating between distant points, the regularization effect of ICT applies to larger regions of the input space.

Interpolation-based regularizers (Tokozume et al., 2018; Verma et al., 2018; Zhang et al., 2018) have been recently proposed for supervised learning, achieving state-of-the-art performances across a variety of tasks and network architectures. While Tokozume et al. (2018) and Zhang et al. (2018) proposed to perform interpolations in the input space, Verma et al. (2018) proposed to perform interpolation also in the hidden space representations. Furthermore, in the unsupervised learning setting, Berthelot, Raffel et al. (2019) proposes to measure the realism of latent space interpolations from an autoencoder to improve its training.

We note that after the publication of an earlier version of this paper (Verma et al., 2019), the methods in Berthelot et al. (2020), Berthelot, Carlini et al. (2019) and Sohn et al. (2020) have achieved state-of-the-art experimental results on benchmark datasets and architectures. Similar to Verma et al. (2019), Berthelot et al. (2020), Berthelot, Carlini et al. (2019) use interpolation in the samples and their predicted targets for designing semi-supervised objective functions. Sohn et al. (2020) combines consistency regularization with pseudo-labeling (Lee, 2013).

Other works have approached semi-supervised learning from the perspective of generative models. Some have approached this from a consistency point of view, such as Lecouat et al. (2018), who proposed to encourage smooth changes to the predictions along the data manifold estimated by the generative model (trained on both labeled and unlabeled samples). Others have used the discriminator from a trained generative adversarial network (Goodfellow et al., 2014) as a way of extracting features for a purely supervised model (Radford et al., 2015). Still, others have used trained inference models as a way of extracting features (Dumoulin et al., 2016).

Our main objective in this paper is to improve *generalization*, instead of the causality invariance. While we mathematically analyzed the *generalization* properties of consistency-regularization with mixup, Han et al. (2021) recently discussed that intuitively, consistency-regularization in general might be able to promote the *causality* invariance too. To understand this, consider a causality invariance such that a causality from an input to a class label is not destroyed under some input perturbation or argumentation. Then, intuitively, consistency-regularization with this particular input perturbation or argumentation might be able to approximately promote the causality invariance. It would be interesting to mathematically formalize this intuition for ICT to quantify the degree of causality invariance that can be promoted by ICT in future work.

## 6. Conclusion

Machine learning is having a transformative impact on diverse areas, yet its application is often limited by the amount of available labeled data. Progress in semi-supervised learning techniques holds promise for those applications where labels are expensive to obtain. In this paper, we have proposed a simple but efficient semi-supervised learning algorithm, Interpolation Consistency Training (ICT), which has two advantages over previous approaches to semi-supervised learning. First, it uses almost no additional computation, as opposed to computing adversarial perturbations or training generative models. Second, it outperforms strong baselines on two benchmark datasets, even without an extensive hyperparameter tuning. Finally, we have shown how ICT can act as a regularizer on the derivatives of all orders and reduce overfitting when confidence values on unlabeled data are high, which can be achieved by additionally using pseudo-labels on the unlabeled data. Our theoretical results predict a failure mode of ICT with low confidence values (without using pseudo-labels), which was confirmed in the experiments, providing a practical guidance to use ICT with high confidence values. As for the future work, extending ICT to interpolations not only at the input but at hidden representations (Verma et al., 2018) could improve the performance even further.

## Declaration of competing interest

The authors declare that they have no known competing financial interests or personal relationships that could have appeared to influence the work reported in this paper.

## Acknowledgments

This work was supported by the Academy of Finland Flagship programme: Finnish Center for Artificial Intelligence (FCAI). We would also like to acknowledge Compute Canada for providing computing resources used in this work.

## Appendix A. Additional lemmas

The following lemmas are used in the proof of our theorems. Their proofs directly follow from those from previous works (Bartlett & Mendelson, 2002; Mohri & Medina, 2014; Mohri et al., 2012).

**Lemma 4.** *Let $\mathcal{F}_{\hat{S}}$ be a set of maps $x \mapsto f(x)$ that depends on an unlabeled dataset $\hat{S}$. Let $q \mapsto \ell(q, y)$ be a $C$-uniformly bounded function for any $q \in \{f(x) : f \in \mathcal{F}_{\hat{S}}, x \in \mathcal{X}\}$ and $y \in \mathcal{Y}$. Then, for any $\delta > 0$, with probability at least $1 - \delta$ over an i.i.d. draw of $m$ samples $((x_i, y_i))_{i=1}^m$, the following holds:*

$$\mathcal{R}_m(\mathcal{F}_{\hat{S}}) \leq \hat{\mathcal{R}}_m(\mathcal{F}_{\hat{S}}) + C\sqrt{\frac{\log(1/\delta)}{m}}.$$

**Proof.** Since changing one point in $S$ changes $\hat{\mathcal{R}}_m(\mathcal{F}_{\hat{S}})$ by at most $C/m$, McDiarmid's inequality implies the statement of this lemma. □

**Lemma 5** (*Talagrand's Contraction Lemma by Ledoux & Talagrand, 2013; Mohri & Medina, 2014*)**.** *Let $\mathcal{F}$ be a set of functions mapping $\mathcal{X}$ to $\mathbb{R}$. Let $\Psi_1, \ldots, \Psi_m$ be $\mu$-Lipschitz functions for some $\mu > 0$. Then, for any samples $S$ of $m$ points $x_1, \ldots, x_m \in \mathcal{X}$, the following inequality holds:*

$$\frac{1}{m}\mathbb{E}_\sigma\left[\sup_{f\in\mathcal{F}}\sum_{i=1}^m \xi_i(\Psi_i \circ f)(x_i)\right] \leq \frac{\mu}{m}\mathbb{E}_\sigma\left[\sup_{f\in\mathcal{F}}\sum_{i=1}^m \xi_i f(x_i)\right].$$

**Proof.** See the proof of Mohri and Medina (2014, Lemma 8). □

## Appendix B. Additional experiments

### B.1. Kullback–Leibler (KL) divergence

Throughout the paper, we use the MSE loss for computing the consistency loss by following the previous studies (Laine & Aila, 2017; Tarvainen & Valpola, 2017). While we found the MSE loss to be sufficient for our problems, one may explore the use of the KL divergence for computing the consistency loss. Figs. 4–5 show the results of replacing the MSE loss with the KL divergence for the experiments reported in Fig. 3 (in the main text). Here, the confidence value is defined by $\frac{1}{n}\sum_{i=1}^n |\frac{1}{2} - f_\theta(u_i)|$ with unlabeled data points $(u_i)_{i=1}^n$. Each line in each subplot shows the decision boundary of the predictor $f_\theta$ (i.e., $\{u : f_\theta(u)\} = \frac{1}{2}$) after each update of 1, 10, 100, and 1000. As we can see in the figures, with the KL divergence (instead of the MSE loss), ICT still works as expected and the numerical results are still consistent with our theoretical prediction regarding the importance of the confidence value. Given these preliminary results, it might be interesting to study the use of the KL divergence in ICT more comprehensively in future work.

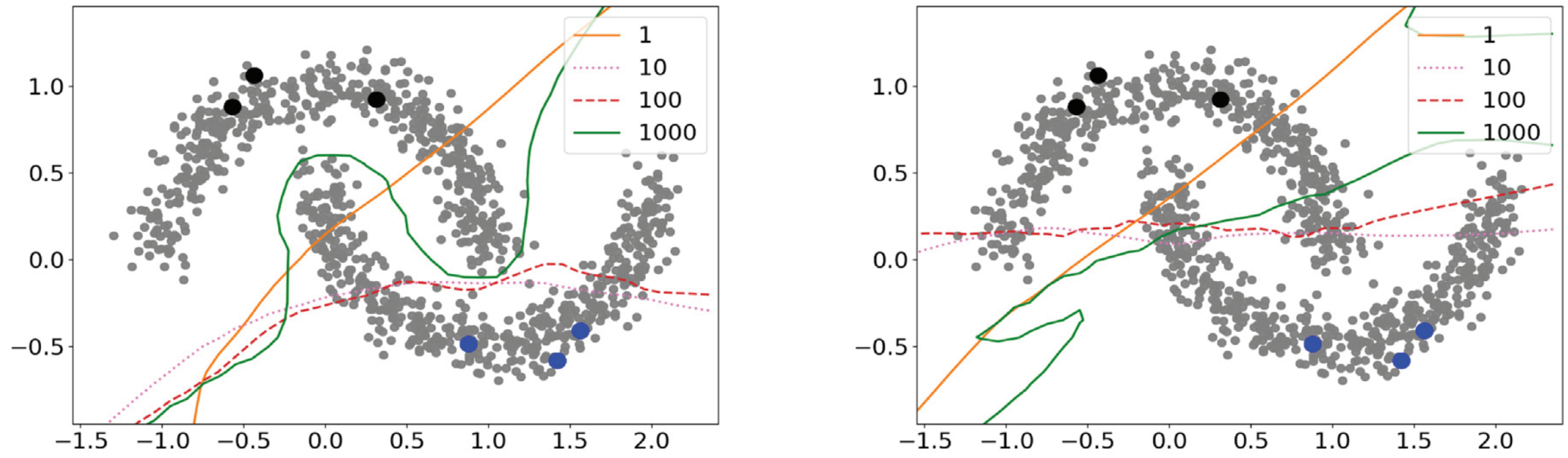

(a) High confidence value = 0.4500 (b) Low confidence value = 0.3115

**Fig. 4.** Decision boundaries for ICT with the KL divergence and the softplus activation.

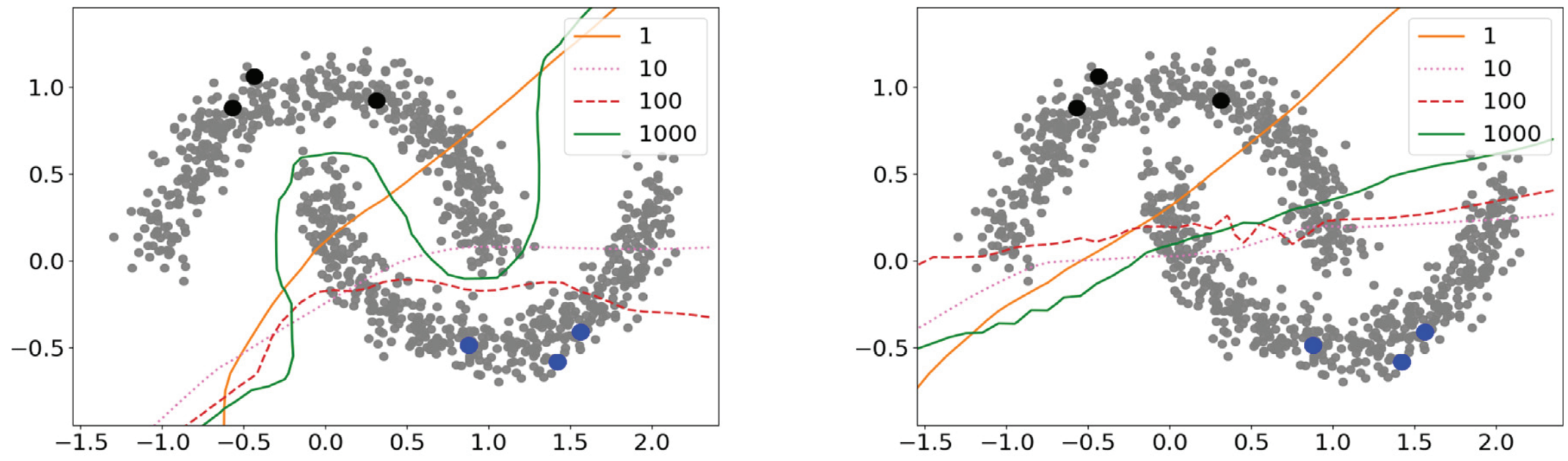

(a) High confidence value = 0.4564 (b) Low confidence value = 0.2966

**Fig. 5.** Decision boundaries for ICT with the KL divergence and the ReLU activation.

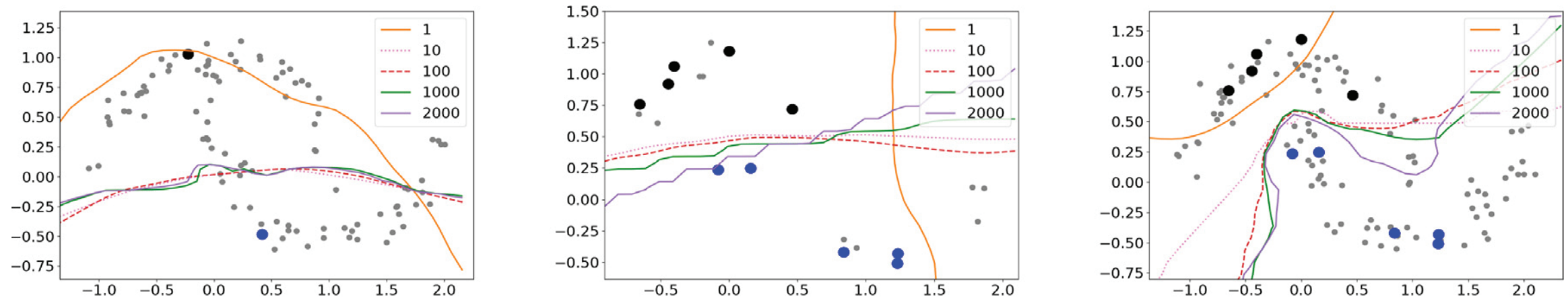

(a) (b) (c)

**Fig. 6.** The effect of few data points on ICT: (a) 1 labeled point per class and 100 unlabeled points, (b) 5 labeled points per class and 10 unlabeled points, and (c) 5 labeled points per class and 100 unlabeled points.

*B.2. The effect of few data points*

As ICT is designed for semi-supervised learning instead of one-shot learning, we now examine how ICT could fail in the case of too few data points in Fig. 6. Each line in each subplot shows the decision boundary of the predictor $f_\theta$ (i.e., $\{u : f_\theta(u)\} = \frac{1}{2}$) after each update of 1, 10, 100, 1000, and 2000. The experiments were conducted with the softplus activation function $\phi(z) = \ln(1 + \exp(\kappa z))/\kappa$ with $\kappa = 100$.

As can be seen in the figure, if we have only one labeled point per class (Fig. 6a) or if unlabeled points are too few (Fig. 6b), then ICT would not learn the correct decision boundary. However, by increasing only the number of *unlabeled* points from 10 (Fig. 6b) to 100 (Fig. 6c), ICT starts learning a good decision boundary (Fig. 6c). The decision boundary is further refined by increasing *only* the unlabeled data points in Figs. 1 and 3 in the main text. Along with our result in Table 2 where ICT performed well only with 250 labeled points for SVHN, these observations further demonstrate the fact that ICT works well with few labeled data points (but not too few such as one per class) as long as we can increase the number of the *unlabeled* data points. This is consistent with our theoretical prediction.

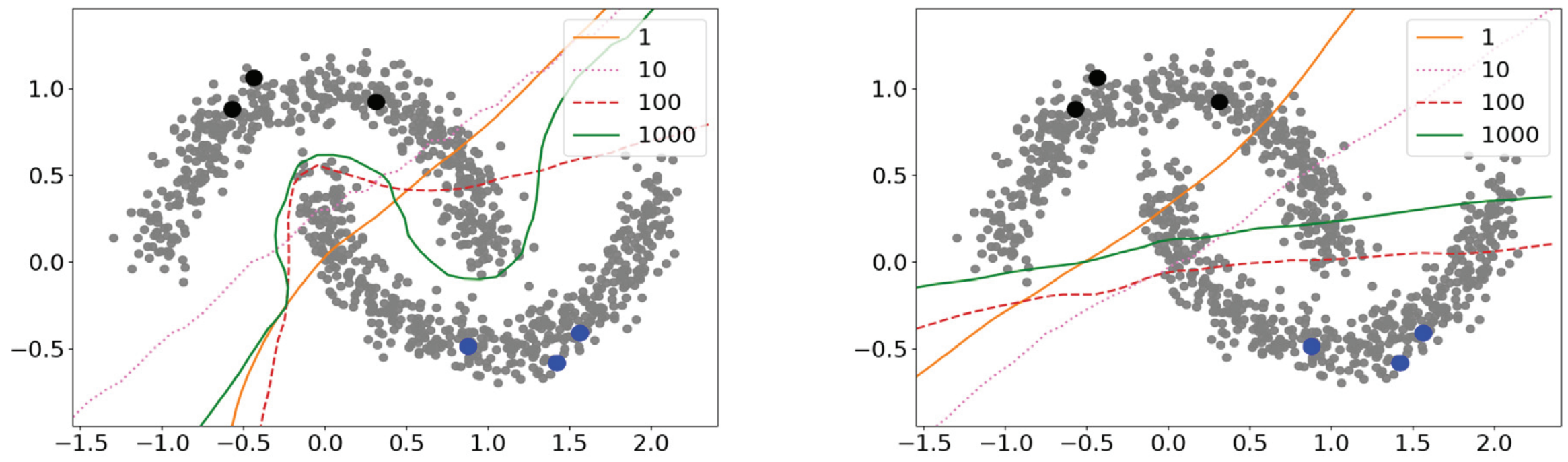

(a) High confidence value = 0.4437 (b) Low confidence value = 0.3026

**Fig. 7.** Decision boundaries for ICT with the ReLU activation (and the MSE loss).

### *B.3. Using the ReLU activation in the experiments of Fig. 3*

In Fig. 3, we used the softplus activation function $\phi(z) = \ln(1 + \exp(\kappa z))/\kappa$ with $\kappa = 100$. As an additional experiment, we replaced the softplus activation by the ReLU activation and reported the results in Fig. 7. As can be seen in the figure, the numerical results with the ReLU activation are also consistent with our theoretical prediction. This is expected because the ReLU function is approximated arbitrarily well with the softplus activation function by varying the value of $\kappa \in (0, \infty)$, and our theoretical results hold for any $\kappa \in (0, \infty)$.